%% file: arxiv.tex
\newif\ifdraft \draftfalse
\newif\iffull \fulltrue

\documentclass[9pt]{article}

\usepackage{booktabs}
\usepackage{subfigure}
\usepackage{algorithm}
\usepackage[noend]{algorithmic}
\usepackage{fullpage}
\usepackage{amsmath, amssymb, amsthm}
\usepackage{mathtools}
\usepackage{color}
\usepackage{xcolor}
\usepackage{natbib}
\usepackage{authblk}

\definecolor{DarkGreen}{rgb}{0.1,0.5,0.1}
\definecolor{DarkRed}{rgb}{0.5,0.1,0.1}
\definecolor{DarkBlue}{rgb}{0.1,0.1,0.5}
\usepackage[]{hyperref}
\hypersetup{
    unicode=false,          
    pdftoolbar=true,        
    pdfmenubar=true,        
    pdffitwindow=false,      
    pdftitle={},    
    pdfauthor={}
    pdfsubject={},   
    pdfnewwindow=true,      
    pdfkeywords={keywords}, 
    colorlinks=true,       
    linkcolor=DarkRed,          
    citecolor=DarkGreen,        
    filecolor=DarkRed,      
    urlcolor=DarkBlue,          
}
\usepackage{xspace}
\usepackage{cleveref}

\newcommand{\sn}[1]{\ifdraft \textcolor{purple}{[Seth: #1]}\fi}

\newcommand\RR{\mathbb{R}}
\newcommand\cA{\mathcal{A}}

\newcommand\cF{\mathcal{G}}
\newcommand\cL{\mathcal{L}}
\newcommand\cG{\mathcal{G}}
\newcommand\cH{\mathcal{H}}

\newcommand\cP{\mathcal{P}}

\newcommand\cX{\mathcal{X}}

\newcommand\knrw{\mathrm{SUBGROUP}}
\newcommand\MSR{\mathrm{MARGINAL}}

\DeclareMathOperator*{\Expectation}{\mathbb{E}}
\newcommand{\Ex}[2]{\Expectation_{#1}\left[#2\right]}

\newcommand\FP{\mathrm{FP}}

\newcommand\LC{\mathrm{LC}}
\newcommand\CSC{\mathrm{CSC}}

\newcommand\fpsize{\alpha_{FP}}
\newcommand\fpdisp{\beta_{FP}}

\def\epsilon{\varepsilon}
\DeclareMathOperator{\OPT}{OPT}
\DeclareMathOperator*{\argmin}{\mathrm{argmin}}
\DeclareMathOperator*{\argmax}{\mathrm{argmax}}
\newcommand{\INDSTATE}[1][1]{\STATE\hspace{#1\algorithmicindent}}

\newtheorem{theorem}{Theorem}[section]

\newtheorem{remark}[theorem]{Remark}

\theoremstyle{definition}
\newtheorem{definition}[theorem]{Definition}

\begin{document}

\title{An Empirical Study of Rich Subgroup\\ Fairness for Machine Learning}


\author[1]{Michael Kearns}
\author[1]{Seth Neel}
\author[1]{Aaron Roth}
\author[2]{Zhiwei Steven Wu}
\affil[1]{University of Pennsylvania}
\affil[2]{University of Minnesota}





\maketitle

\begin{abstract}
  \cite{KNRW18} recently proposed a notion of \emph{rich subgroup
    fairness} intended to bridge the gap between statistical and
  individual notions of fairness. Rich subgroup fairness picks a
  statistical fairness constraint (say, equalizing false positive
  rates across protected groups), but then asks that this constraint
  hold over an exponentially or infinitely large collection of
  \emph{subgroups} defined by a class of functions with bounded VC
  dimension.  They give an algorithm guaranteed to learn subject to
  this constraint, under the condition that it has access to oracles
  for perfectly learning absent a fairness constraint. In this paper,
  we undertake an extensive empirical evaluation of the algorithm of
  Kearns et al.  On four real datasets for which fairness is a
  concern, we investigate the basic convergence of the algorithm when
  instantiated with fast heuristics in place of learning oracles,
  measure the tradeoffs between fairness and accuracy, and compare
  this approach with the recent algorithm of \cite{MSR}, which
  implements weaker and more traditional marginal fairness constraints
  defined by individual protected attributes. We find that in general,
  the Kearns et al.  algorithm converges quickly, large gains in
  fairness can be obtained with mild costs to accuracy, and that
  optimizing accuracy subject only to marginal fairness leads to
  classifiers with substantial subgroup unfairness.  We also provide a
  number of analyses and visualizations of the dynamics and behavior
  of the Kearns et al. algorithm.  Overall we find this algorithm to
  be effective on real data, and rich subgroup fairness to be a viable
  notion in practice.
\end{abstract}

\input{intro}

\input{defs}

\input{empirical}

\input{conclusions}

\bibliographystyle{plainnat}
 \bibliography{./arxiv.bbl}

 \clearpage
 \newpage

 \appendix
 \input{appendix}




\end{document}

%% file: intro.tex
\section{Introduction}

The most common definitions of fairness in machine learning are statistical in nature. 
They proceed by fixing a small number of ``protected subgroups'' (such as racial or gender groups), and then ask that some statistic of interest be approximately equalized across groups. Standard choices for these statistics include positive classification rates \citep{CV10}, false positive or false negative rates \citep{HPS16,KMR16,Chou16} and positive predictive value \citep{Chou16,KMR16} --- see \cite{Berksurvey} for more examples. These definitions are pervasive in large part because they are easy to check, although there are interesting computational challenges in learning subject to these constraints in the worst case --- see e.g. \cite{hardnesspaper}.

\begin{figure}
\centering
\includegraphics[scale=0.3]{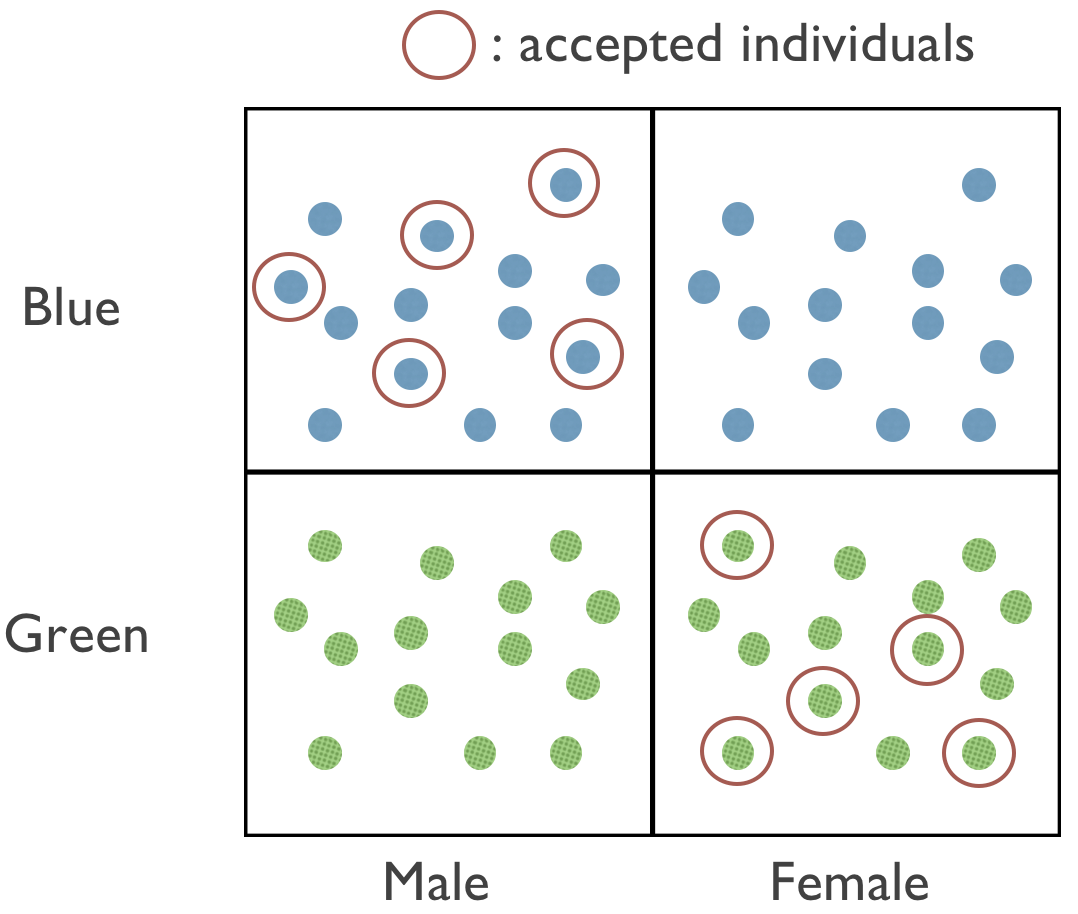}
\caption{Fairness Gerrymandering: A Toy Example \citep{KNRW18}}
\label{fig:example}
\end{figure}

 Unfortunately, these statistical definitions are not very meaningful to individuals: because they are constraints only over \emph{averages} taken over large populations, they promise essentially nothing about how an individual person will be treated. \citet{awareness} enumerate a ``catalogue of evils'' which show how definitions of this sort can fail to provide meaningful guarantees. \citet{KNRW18} identify a particularly troubling failure of standard statistical definitions of fairness, which can arise naturally without malicious intent, called ``fairness gerrymandering''. They illustrate the idea with the following toy example shown in Figure \ref{fig:example}, described as follows.

Suppose individuals each have two sensitive attributes: race (say blue and green) and gender (say male and female). Suppose that these two attributes are distributed independently and uniformly at random, and are uncorrelated with a binary label that is also distributed uniformly at random. If we view gender and race as defining classes of people that we wish to protect, we could take a standard statistical fairness definition from the literature --- say the equal odds condition of \cite{HPS16}, which asks to equalize false positive rates across protected groups, and instantiate it with the four protected groups: ``Men'', ``Women'', ``blue people'', and ``green people''. The following classifier satisfies this condition, although only by ``cheating'' and packing its unfairness into structured subgroups of the protected populations: it labels a person as positive only if they are a blue man or a green woman. This equalizes false positive rates across the four specified groups, but of course not over the finer-grained subgroups defined by the intersections of the two protected attributes.

\citet{KNRW18} also proposed an approach to the problem of fairness
gerrymandering: rather than asking for statistical definitions of
fairness that hold over a small number of coarsely defined groups, ask
for them to hold over a combinatorially or infinitely large collection
of subgroups defined by a set of functions $\cF$ of the protected
attributes (\citet{multical} independently made a similar
proposal). For example, we could ask to equalize false positive rates
across every subgroup that can be defined as the intersection or
conjunction of $d$ protected attributes, for which there are $2^d$
such groups. \cite{KNRW18} showed that as long as the class of
functions defining these subgroups has bounded VC dimension, the
statistical learning problem of finding the best (distribution over)
classifiers in $\cH$ subject to the constraint of equalizing the
positive classification rate, the false positive rate, or the false
negative rate over every subgroup defined over $\cF$ is solvable
whenever the dataset size is sufficiently large relative to the VC
dimension of $\cF$ and $\cH$. Taking inspiration from the technique of
\citet{MSR}, they were able to show that even with
combinatorially many subgroup fairness constraints, the computational
problem of learning the optimal fair classifier is once again solvable
efficiently whenever the learner has access to a black-box classifier
(oracle) which can solve the \emph{unconstrained} learning problems
over $\cF$ and $\cH$ respectively. Similarly, given access to an
oracle for $\cF$, they were able to efficiently solve the problem of
\textit{auditing} for rich subgroup fairness: finding the $g \in \cF$
that corresponds to the subgroup for whom the statistical fairness
constraint was most violated.

While the work of \citet{KNRW18} is satisfying from a theocratical point of view, it leaves open a number of pressing empirical questions. For example, their theory is built for an idealized setting with perfect learning oracles --- in practice heuristic oracles may fail. Moreover, perhaps rich subgroup fairness is asking for too much in practice --- maybe enforcing combinatorially many constraints leads to an untenable tradeoff with error.  Finally, perhaps enforcing combinatorially many constraints is not necessary --- perhaps on real data, it is enough to call upon the algorithm of \cite{MSR} for enforcing statistical fairness constraints on the small number of groups defined by the marginal protected attributes, and rich subgroup fairness will follow incidentally. Put another way: Is the so-called \textit{fairness gerrymandering} problem only a theoretical curiosity, or does it arise organically when standard classifiers are optimized subject to marginal statistical fairness constraints?

In this paper, we conduct an extensive set of experiments to answer these questions. We study the algorithm from \cite{KNRW18} --- instantiated with fast heuristic learning oracles --- when used to train a linear classifier subject to approximately equalizing false positive rates across a rich set of subgroups defined by linear threshold functions. On four real datasets, we characterize:


\begin{enumerate}
\item The basic convergence properties of the algorithm --- although this algorithm has provable guarantees when instantiated with learning \emph{oracles} for $\cF$ and $\cH$, when these oracles are (necessarily) replaced with heuristics, the guarantees of the algorithm become heuristic as well. We find that the algorithm typically converges (Subsection~\ref{subsec:conv}), and provides a controllable trade-off between fairness and accuracy despite its heuristic guarantees (Subsection~\ref{subsec:Pareto}). We visualize the optimization trajectory of the algorithm (Subsection~\ref{subsec:traj}), and \textit{discrimination heatmaps} showing the evolution of the subgroup discrimination of the algorithm over time (Subsection~\ref{subsec:heatmap}).
    \item The trade-off between subgroup fairness and accuracy. We find that for each dataset, there are appealing compromises between
	error and subgroup fairness. Thus achieving rich subgroup fairness may be possible in practice
	without a severe loss in predictive accuracy (Subsection~\ref{subsec:Pareto}).
\item The subgroup (unfairness) that can result when one applies more standard approaches, that either ignore fairness constraints all together, or equalize false positive rates only across a small number of subgroups defined by individual protected attributes. By \textit{auditing} the models produced by these standard approaches with the rich subgroup auditor of \cite{KNRW18}, we find that often subgroup fairness constraints are violated, even by algorithms which are explicitly equalizing false positive rates across the groups defined on the marginal protected attributes.
\end{enumerate}

In light of these findings, we submit that rich subgroup fairness constraints are both important, and can be satisfied at reasonable cost: both in terms of computation, and in terms of accuracy. We hope that algorithms like that of \cite{KNRW18} which can be used to satisfy rich subgroup fairness become part of the standard toolkit for fair machine learning. 

\subsection{Further Related Work}
While \citet{KNRW18} propose and study rich sub-group fairness for false positive and negative constraints, \citet{multical} study the analogous notion for \emph{calibration} constraints, which they call \emph{multi-calibration}. \citet{multiac} extend this style of analysis to \emph{accuracy} constraints (asking that a classifier be equally accurate on a combinatorially large collection of subgroups). \citet{multimet} also extend it to metric fairness constraints, converting the \emph{individual} metric fairness constraint of \citet{awareness} into a statistical constraint that asks that \emph{on average}, individuals in (combinatorially many) subgroups should be treated differently only in proportion to the average difference between individuals in the subgroups, as measured with respect to some similarity metric.

%
%
%
%
%

%% file: defs.tex
\section{Definitions}
We begin with some definitions, following the notation in
\cite{KNRW18}.  We study the classification of individuals defined by
a tuple $((x, x'), y)$, where $x\in \cX$ denotes a vector of
\emph{protected attributes}, $x'\in \cX'$ denotes a vector of
\emph{unprotected attributes}, and $y\in \{0, 1\}$ denotes a label. We
will write $X = (x, x')$ to denote the joint feature vector. We assume
that points $(X, y)$ are drawn i.i.d. from an unknown distribution
$\mathcal{P}$.  Let $D$ be a binary classifier, and let
$D(X) \in \{0,1\}$ denote the (possibly randomized) classification
induced by $D$ on individual $(X,y)$.


We will be concerned with learning and auditing classifiers $D$
satisfying a common statistical fairness constraint: equality
of false positive rates (also known as equal opportunity). The techniques in \citet{MSR} and \citet{KNRW18} also apply equally well to equality of false negative rates and equality of
classification rates (also known as statistical parity).\footnote{or more generally to any fairness constraint  that can be expressed as a linear equality on the conditional moments $\Ex{}{t(X, y, D(X) | \epsilon(X,y))},$ where $\epsilon(X,y)$ is an event defined with respect to $(X, y)$, and
$t: X \times \{0,1\} \times \{0,1\} \to [0,1]$ \cite{MSR}.  Equality of false positive rate is a particular instantiation of this kind of constraint where $\epsilon$  is the event $y = 0$, and $t = \textbf{1}\{D(X) =1\}$.}

Each fairness constraint is defined with
respect to a set of protected groups. We define sets of protected
groups via a family of indicator functions $\cF$ for those groups,
defined over protected attributes. Each $g:\cX \to \{0,1\} \in \cF$
has the semantics that $g(x) = 1$ indicates that an individual with
protected features $x$ is in group $g$.
 We now formally define
false positive subgroup fairness.

\begin{definition}[False Positive Subgroup Fairness]\label{fp-fair}
  Fix any classifier $D$, distribution $\mathcal{P}$, collection of
  group indicators $\cF$, and parameter $\gamma \in [0,1]$.  For each
  $g\in \cG$, define
\begin{align*}
  \fpsize(g, \cP) = \Pr_{\cP}[g(x) = 1, y=0],\\
  \fpdisp(g, D, \cP) = \left| \FP(D) - \FP(D, g) \right|
\end{align*}
  where $\FP(D) = \Pr_{D, \cP} [D(X) = 1\mid y=0]$ and
  $\FP(D, g) = \Pr_{D, \cP}[D(X) = 1\mid g(x) =1 , y=0]$ denote the
  overall false-positive rate of $D$ and the false-positive rate of
  $D$ on group $g$ respectively.

  We say $D$
  satisfies $\gamma$-\emph{False Positive (FP) Fairness} with respect
  to $\mathcal{P}$ and $\cF$ if for every $g \in \cG$
  \[ \fpsize(g, \cP) \cdot \fpdisp(g, D, \cP) \leq \gamma.\] We will
  sometimes refer to $\FP(D)$ FP-base rate.
\end{definition}

Since we do not consider other measures in this paper, we refer to this notion as simply ``subgroup fairness.'' Given a fixed subgroup $g \in \cG$ we will refer to the quantity $\fpsize(g, \cP) \cdot \fpdisp(g, D, \cP)$ as the subgroup fairness wrt $g$, or alternately the $\gamma$-unfairness of $g$. The notion of subgroup fairness imposes a statistical constraint on combinatorially many groups definable by the protected attributes. This is in contrast to more common statistical fairness definitions, defined on coarse groups definable by a single protected attribute. Given a protected attribute $x_i$ and a value for that attribute $a$, define the function $g_{i,a}(x) = \textbf{1}\{x_i = a\}$ denoting the set of individuals who have that particular value of their protected attribute. In contrast to subgroup fairness, we refer to a classifier $D$ as
\textit{marginally} fair if it satisfies false positive subgroup fairness with respect to the functions $\{g_{i,a}\}$ for each protected attribute $x_i$ and realization $a$.

 If the
algorithm $D$ fails to satisfy the $\gamma$-subgroup fairness condition, then
we say that $D$ is $\gamma$-\emph{unfair} with respect to
$\mathcal{P}$ and $\cF$. We call any subgroup $g$ which witnesses this
unfairness a $\gamma$-\emph{unfair certificate} for
$(D, \mathcal{P})$.

An \emph{auditing algorithm} for a notion of fairness is given sample access to points from the underlying distribution, as well as the classification outcomes provided by $D$. It will either deem $D$ to be fair with respect to $\mathcal{P}$, or else produces a certificate of unfairness.
%
%

The algorithms of \citet{MSR} and \citet{KNRW18} studied in this paper both assume access to oracles which can solve \emph{cost-sensitive classification
  (CSC)} problems. Formally, an instance of a CSC problem for the
class $\cH$ is given by a set of $n$ tuples
$\{(X_i, c_i^0, c_i^1)\}_{i=1}^n$ such that $c_i^\ell$ corresponds to
the cost for predicting label $\ell$ on point $X_i$. Given such an
instance as input, a CSC oracle finds a hypothesis $\hat h \in \cH$ that
minimizes the total cost across all points:
\begin{equation}
  \hat h \in \argmin_{h\in \cH} \sum_{i = 1}^n [h(X_i) c_i^1 + (1 -
  h(X_i)) c_i^0]\label{csc}
\end{equation}

Following both \cite{MSR} and \citet{KNRW18}, in all of the experiments in this paper we take the classes $\cH$ and $\cG$ to be linear threshold functions, and we use a linear regression heuristic for both auditing and learning. The heuristic finds a linear threshold function as follows: \\
\begin{itemize}
\item Train two linear regression models $r_0$, $r_1$ to predict $c_0$ and $c_1$ respectively.
\item Given a new point $x$, predict
the cost of classifying $x$ as $0$ and $1$ using our regression models: these are $r_0(x)$ and
$r_1(x)$ respectively.
\item Output the prediction $\hat{y}$ corresponding to lower predicted cost:
$\hat{y} = \argmin_{i \in \{0,1\}}r_i(x)$.
\end{itemize}
We leave the precise descriptions of the algorithm from \cite{KNRW18}
--- which we will refer to as the $\knrw$ algorithm --- to the
appendix. We refer the reader to \cite{KNRW18} for details about its
derivation and guarantees.\footnote{\cite{KNRW18}
  actually give two algorithms, one of which employs no-regret learning techniques and converges in a polynomial
  number of rounds, but is randomized; and the other of which
  is known to converge only in the limit (but is conjectured to converge
  quickly), and is deterministic. We focus on
  the deterministic algorithm in this paper, because it is more
  amenable to implementation, despite its weaker theoretical
  guarantees. We find that it performs well in practice despite its 
  weaker theory.}  At this point we
remark only that the algorithm operates by expressing the optimization
problem to be solved (minimize error, subject to subgroup fairness
constraints) as solving for the equilibrium in a two player zero-sum
game, between a \emph{Learner} and an \emph{Auditor}. The
\emph{Learner} has the set of hypothesis $\cH$ as its action (pure strategy) space,
and the \emph{Auditor} has the set of subgroups $\cF$ as its action
space. The best response problem for the Auditor corresponds to the
auditing problem: finding the subgroup $g \in \cF$ for which the
strategy of the learner violates the fairness constraints the
most. The best response problem for the Learner corresponds to solving
a weighted (but unconstrained) empirical risk minimization
problem. The best response problem for both players can be expressed
as solving a cost sensitive classification problem. The algorithm
$\knrw$ essentially simulates the \emph{fictitious play} of this game,
which proceeds over rounds, and in each round $t$ both players best
respond to their opponent's empirical history of play:
\begin{itemize} 
\item Learner plays $h_t$ in $\cH$ that minimizes objective function
  balancing error and unfairness on subgroups $g_1, \ldots, g_{t-1}$
  found by Auditor so far;
\item Auditor finds subgroup $g_t$ in $\cG$ on which the uniform
  distribution over $h_1, \ldots , h_t$ violates $\gamma$-fairness the
  most.
\end{itemize}
 This can be done efficiently assuming access to oracles which solve the cost sensitive classification problem over $\cF$ and $\cH$ respectively.


%% file: empirical.tex
\begin{table*}[t]
\centering
\begin{tiny}
\begin{tabular}{|l|l|l|l|l|l|l|}
\hline
{\bf Dataset}	& {\bf Size} & {\bf Prediction} & {\bf \#Features} & {\bf \#Protected} & {\bf Protected Feature Types} & {\bf Baseline}\\ \hline
Communities and Crime  & 1994 & High Violent Crime ? & 128 & 18 & Race & 0.3 \\ \hline
Law School	       & 2053 & Pass Bar Exam ? & 10 & 4 & Race, Income, Age, Gender & 0.49 \\ \hline
Student		       & 396  & Course Performance ? & 30 & 5 & Age, Gender, Relationship, Alcohol Use & 0.47 \\ \hline
Adult		       & 2021 & Income $>=$ \$50K ? & 14 & 3 & Age, Race, Gender & 0.50 \\ \hline
\end{tabular}
\end{tiny}
\caption{Description of Data Sets.}
\label{tab:datasets}
\end{table*}

\section{Empirical Evaluation}

In this section, we describe an extensive empirical investigation of the $\knrw$ algorithm on
four datasets in which fairness is a potential concern.
Among the questions of primary interest are the following:
\begin{itemize}
\item Does the $\knrw$ algorithm work in practice, despite the use of imperfect heuristics for the Learner and Auditor?
\item Is the notion of subgroup fairness interesting empirically, in that there are palatable trade-offs between accuracy
	and subgroup fairness (as opposed to it
	being too strong a constraint, and thus resulting in a very steep error increase for even weak subgroup fairness)?
\end{itemize}
We will answer these questions strongly in the affirmative, which is perhaps the overarching message of our results.
We also carefully compare subgroup fairness to standard marginal fairness, and show that optimizing
	for the latter in general does poorly on the former --- thus something like the $\knrw$ algorithm is actually necessary
	to achieve subgroup fairness.

More generally, aside from performance, we provide a number of empirical analyses that elucidate
	the underlying behavior and convergence properties of the $\knrw$ algorithm, and discuss its strengths and
	weaknesses.

\subsection{Datasets}

We ran experiments on $3$ datasets from the UCI Machine Learning
Repository \cite{UCI}:
\textbf{Communities and Crime} \citep{communities},
\textbf{Adult},
and \textbf{Student} \citep{student}, and the \textbf{Law School} dataset from the Law School Admission Council's National Longitudinal Bar Passage Study \citep{lawschool}.
These datasets were selected due to their potential fairness concerns, including:
\begin{itemize}
\item Data points representing individual people (or in the case of Communities and Crimes,
	small U.S. communities of people);
\item The presence of features capturing properties often associated with possible
	discrimination, including race, gender, and age;
\item Potential sensitivity of the predictions being made, such as violent crime,
	income, or performance in school.
\end{itemize}

The properties of these datasets are summarized in Table~\ref{tab:datasets},
including the number of instances, the prediction being made, the overall number of
features (which varies from 10 to 128),
the number of protected features in the subgroup class (which varies from 3 to 18),
the nature of the protected
features, and the baseline (majority class) error rate.

Some methodological notes:
\begin{itemize}
\item We note that two of the datasets (Law School and Adult) were initially much larger
but were extremely imbalanced with respect to the predicted label, making sensible
error comparisons numerically difficult. We thus randomly downsampled these two datasets to
obtain approximately balanced prediction problems on each.
\item All categorical variables have been preprocessed with a one-hot encoding.\item The $\knrw$ algorithm has two input parameters: the maximum allowed subgroup fairness violation.
	$\gamma$, and a tuning parameter $C$ which represents (in the theoretical derivation in \cite{KNRW18}) an upper bound on the magnitude of the dual variables needed to express the fairness constrained empirical risk minimization problem.
	We view $\gamma$ as an important control variable allowing us to explore the tradeoff between
	fairness and accuracy, and thus will vary it in our experiments. On the other hand, $C$ is more of a
	nuisance parameter, and thus for consistency and simplicity we set $C = 10$ in all experiments.
	Experimentation with larger values of $C$ did not reveal qualitatively different findings on the
	datasets investigated.
\item We emphasize that {\em all results are reported in-sample on the datasets\/}, and thus we are treating
	the empirical distributions of the datasets as the ``true'' distributions of interest. We do this
	because our primary interest is simply in examining the performance and behavior of the $\knrw$ algorithm
	on the actual data or distributions, and not in generalization per se. As noted in~\cite{KNRW18},
	theoretical generalization bounds for both error and subgroup fairness
	can be obtained by standard methods, and will depend on (e.g.) the
	VC dimension of the Learner's model class $\cH$ and the Auditor's subgroup class $\cF$. As usual,
	we would expect empirical generalization to often be considerably better than the worst-case theory.
\end{itemize}

\subsection{Empirical Convergence of $\knrw$}
\label{subsec:conv}

\begin{center}
\begin{figure*}[h]
\centering
\subfigure[]{\includegraphics[scale=0.2]{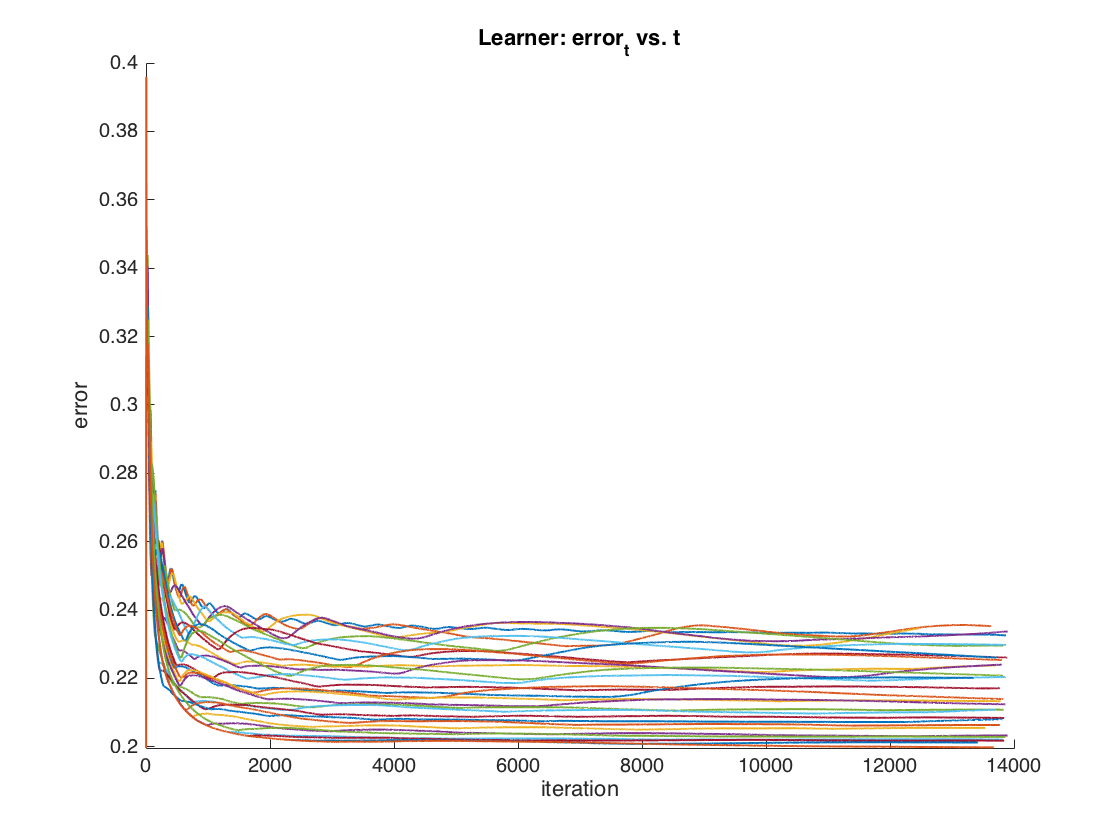}}
\subfigure[]{\includegraphics[scale=0.2]{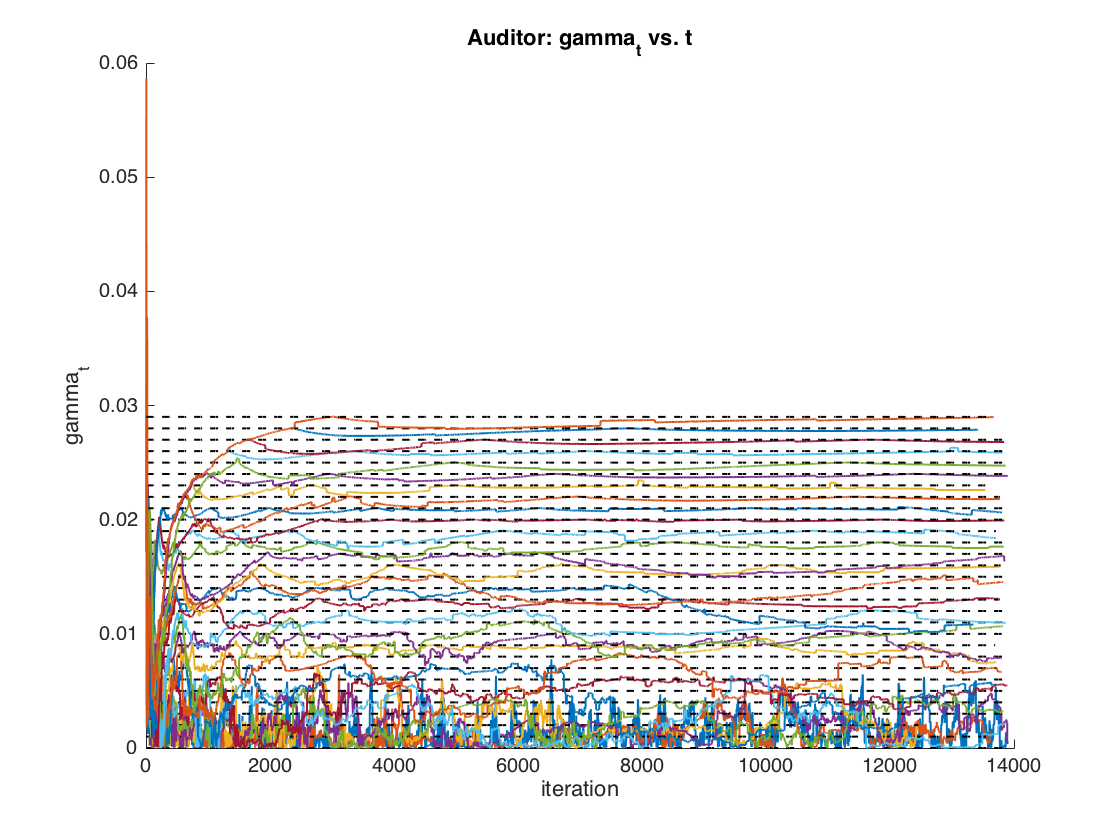}}
\subfigure[]{\includegraphics[scale=0.2]{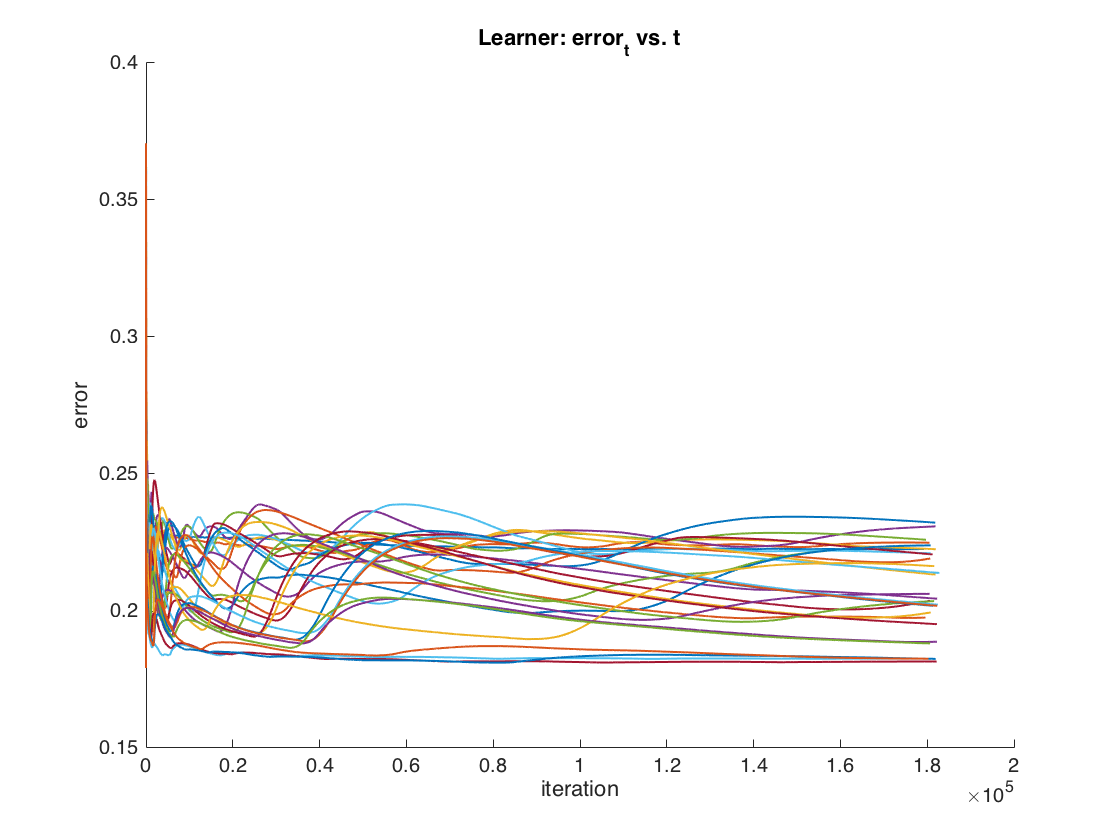}}
\subfigure[]{\includegraphics[scale=0.2]{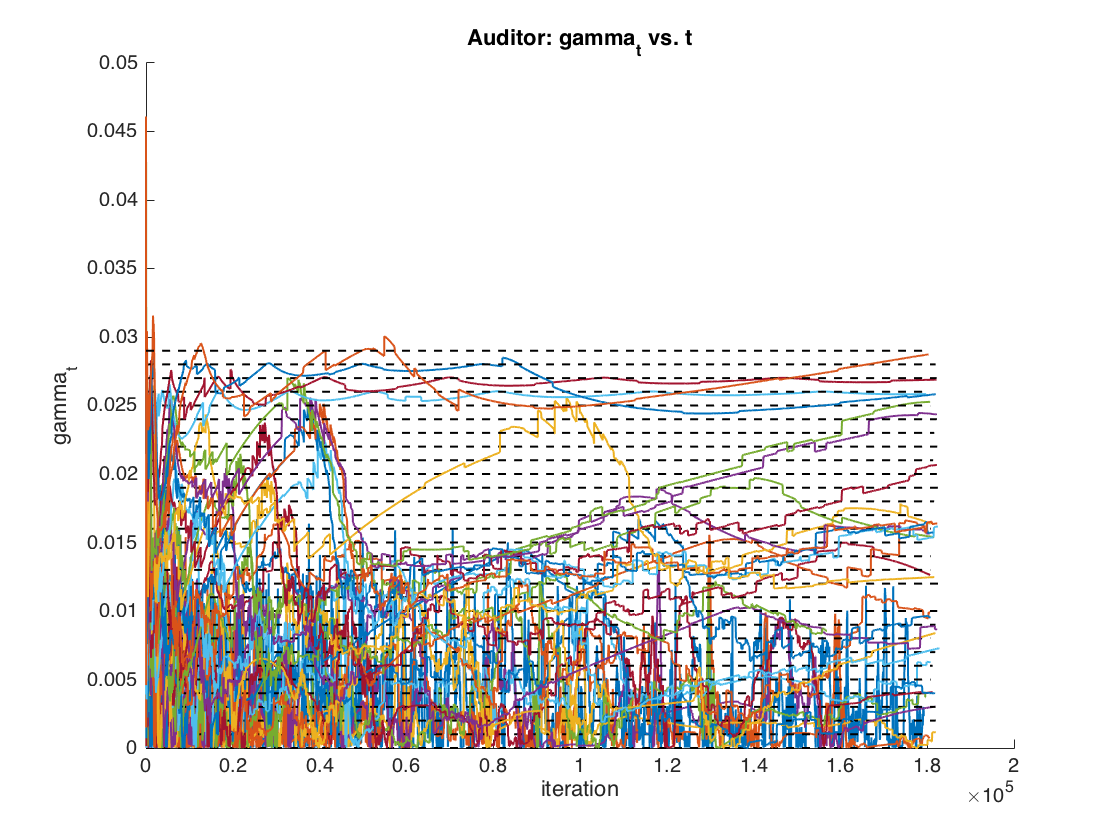}}
\caption{{Error $\epsilon_t$ and fairness violation $\gamma_t$ for Law School dataset (panels (a) and (b)) and Adult data set
(panels (c) and (d)), for values of input $\gamma$ ranging from 0 to 0.03.
Dashed horizontal lines on $\gamma_t$ plots correspond to varying values of $\gamma$.
}}
\label{fig:conv}
\end{figure*}
\end{center}

We begin with an examination of the convergence properties of the $\knrw$ algorithm on the four datasets.
\citet{KNRW18} had already reported preliminary convergence results for the Communities and Crime dataset,
showing that their algorithm converges quickly, and that varying the input $\gamma$ provides an appealing trade-off between
error and fairness.
In addition to replicating those findings for Communities and Crime,
we also find that they are not an optimistic anomaly.
For example, for the Law School dataset, in Figure~\ref{fig:conv} 
we plot both the error $\epsilon_t$ (panel (a)) and the fairness violation
$\gamma_t$ (panel (b)) as a function of the iteration $t$, for values of the input $\gamma$ ranging from 0 to 0.03.
We see that the algorithm converges relatively quickly (on the
order of thousands of iterations), and that
increasing the input $\gamma$ generally yields decreasing error and increasing fairness violation (typically saturating
the input $\gamma$), as suggested
by the idealized theory.

But on other datasets the empirical convergence does not match the idealized theory as cleanly, presumably due to the
use of imperfect Learner and Auditor heuristics. In panels (c) and (d) of Figure~\ref{fig:conv} we again plot $\epsilon_t$
and $\gamma_t$, but now for the Adult dataset. Even after approximately 180,000 iterations, the algorithm does not appear
to have converged, with $\epsilon_t$ still showing long-term oscillatory behavior, $\gamma_t$ exhibiting extremely noisy
dynamics (especially at smaller input $\gamma$ values), and there being no clear systematic, monotonic relationship between
the input $\gamma$ and error acheived. But despite this departure from the theory, it remains the case that varying $\gamma$
still yields a diverse set of $\langle \epsilon_t, \gamma_t \rangle$ pairs, as we will see in the next section.
In this sense, even in the absence of convergence
the algorithm can be viewed as a valuable search tool for models trading off accuracy and fairness.

Overall, we found rather similar convergent behavior on the Communities and Crime and Law School datasets, and
less convergent behavior on the Adult and Student datasets.

\subsection{Subgroup Pareto Frontiers and Comparison to Marginal Fairness}
\label{subsec:Pareto}

\begin{figure*}[p]
\centering
\subfigure[]{\includegraphics[scale=0.14]{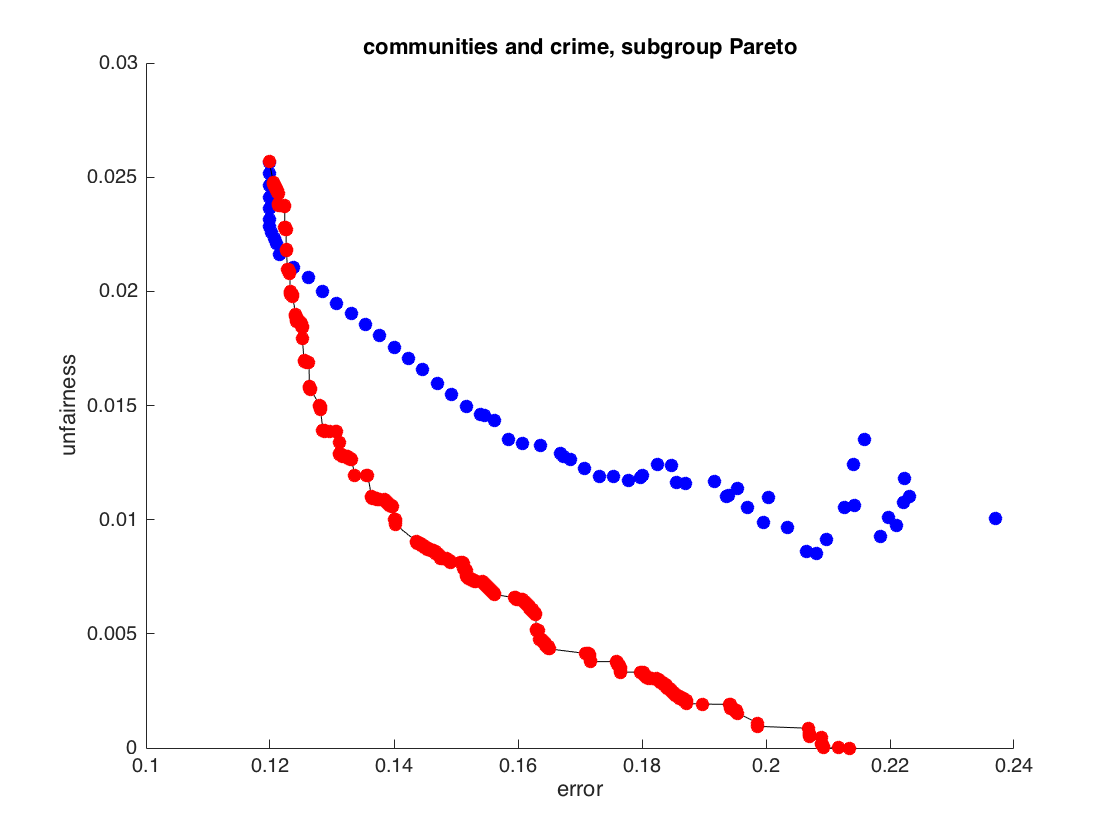}}
\subfigure[]{\includegraphics[scale=0.14]{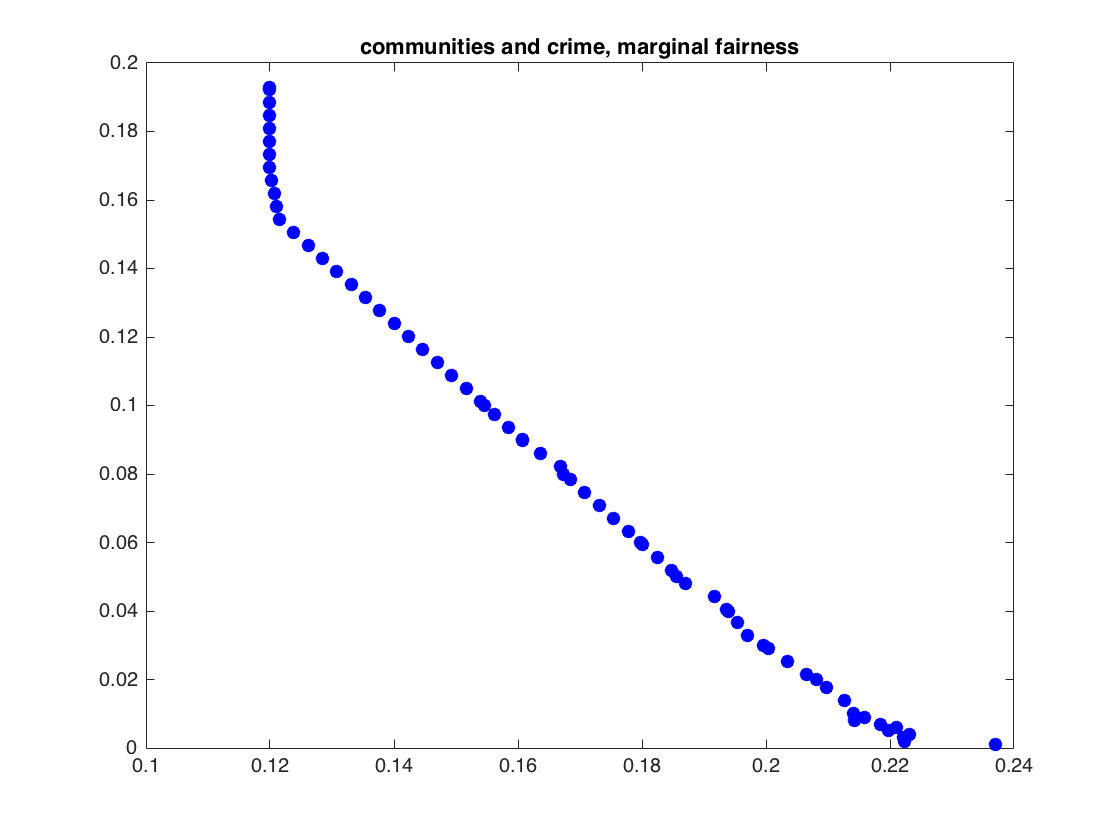}}
\subfigure[]{\includegraphics[scale=0.14]{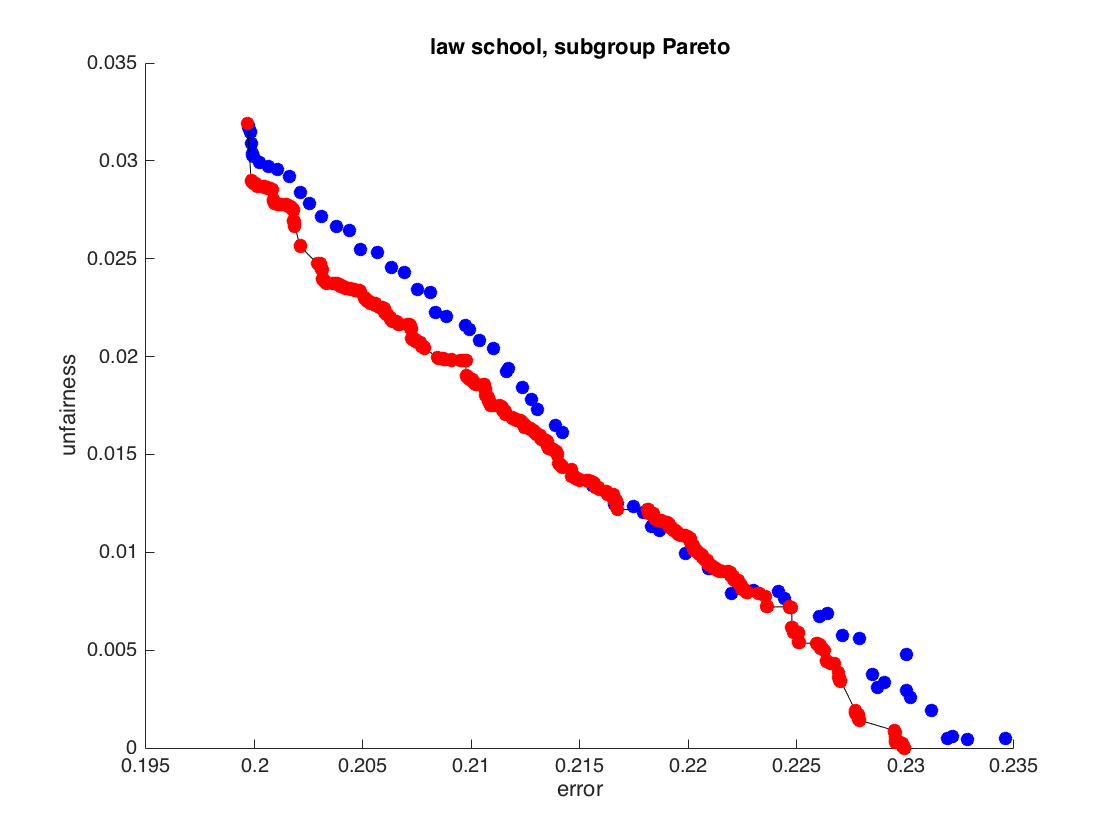}}
\subfigure[]{\includegraphics[scale=0.14]{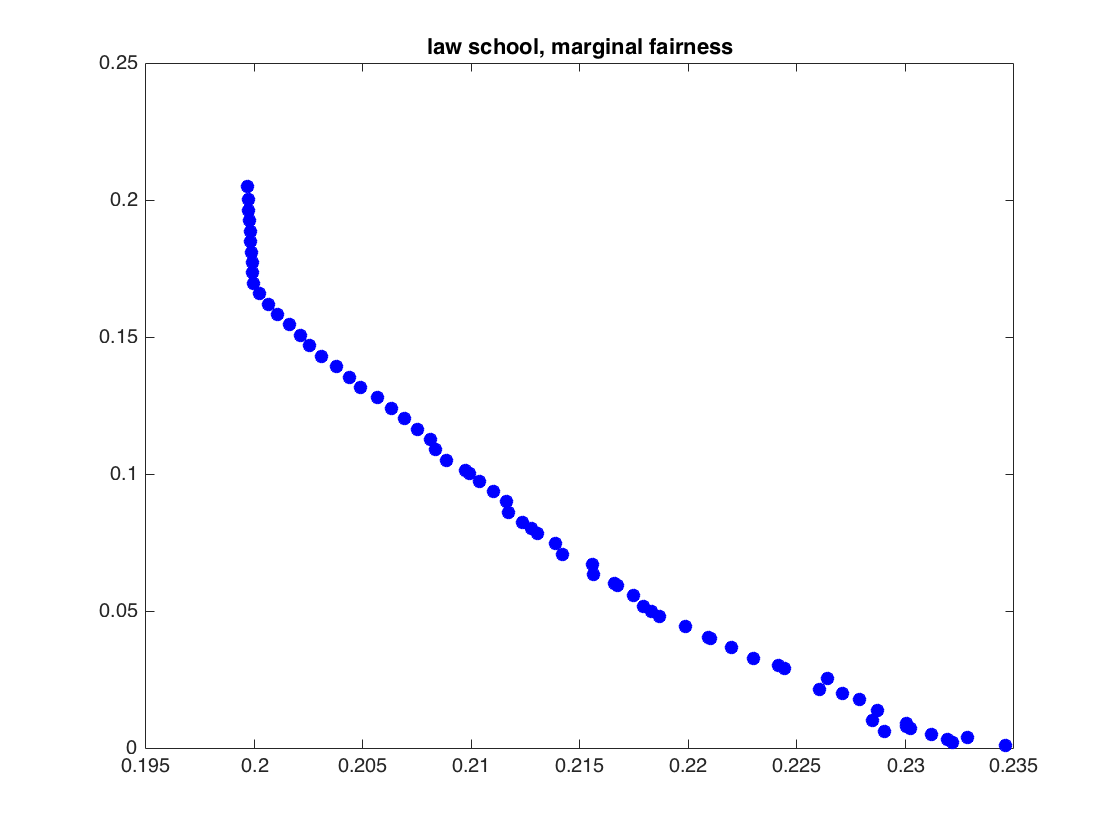}}
\subfigure[]{\includegraphics[scale=0.14]{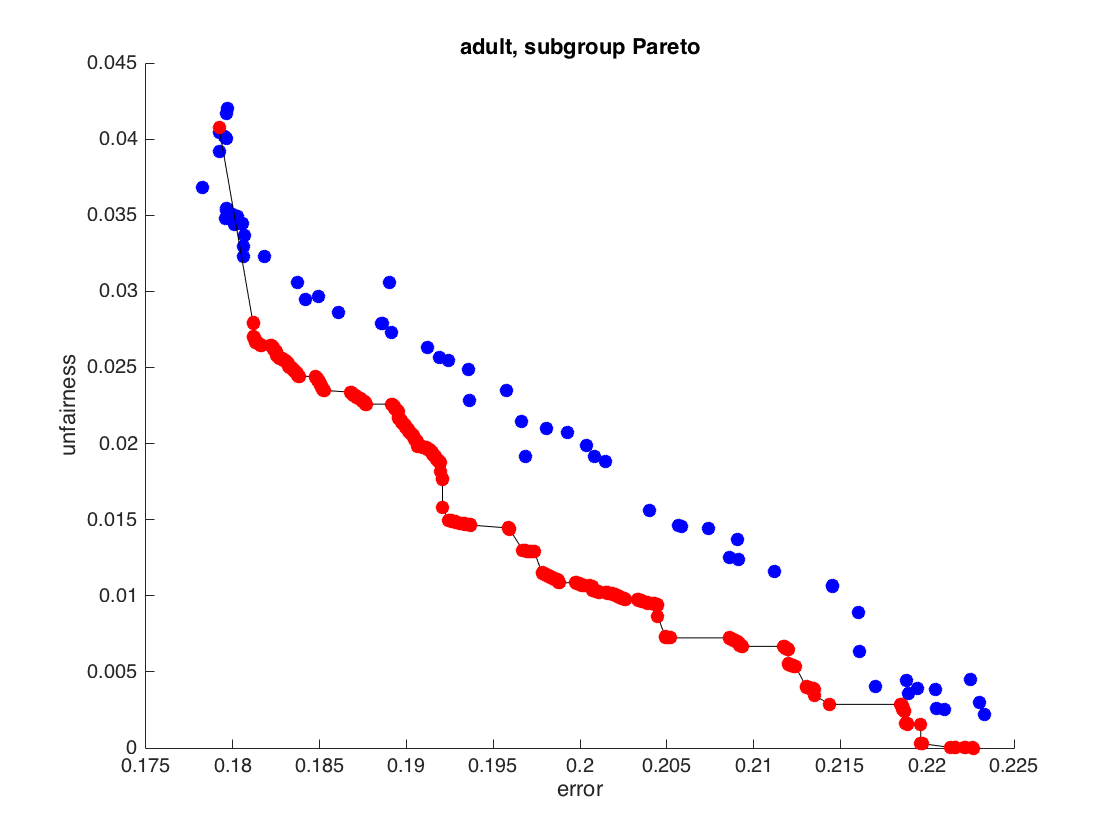}}
\subfigure[]{\includegraphics[scale=0.14]{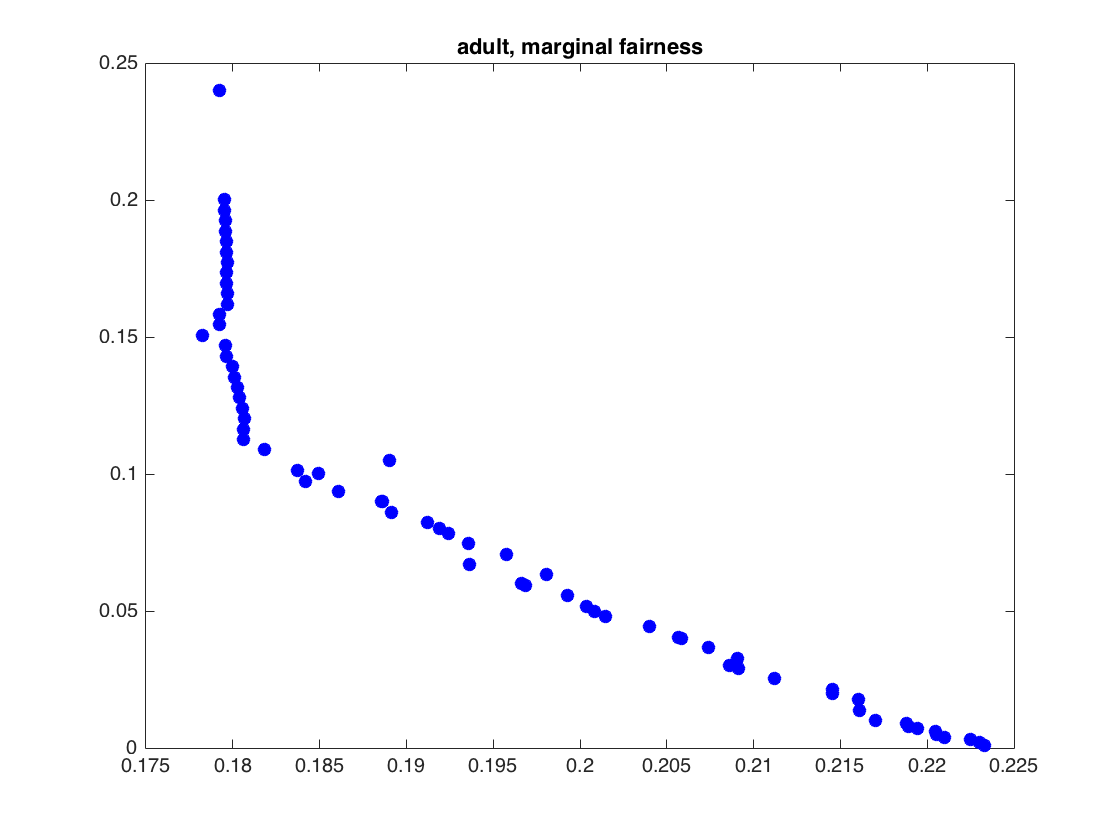}}
\subfigure[]{\includegraphics[scale=0.14]{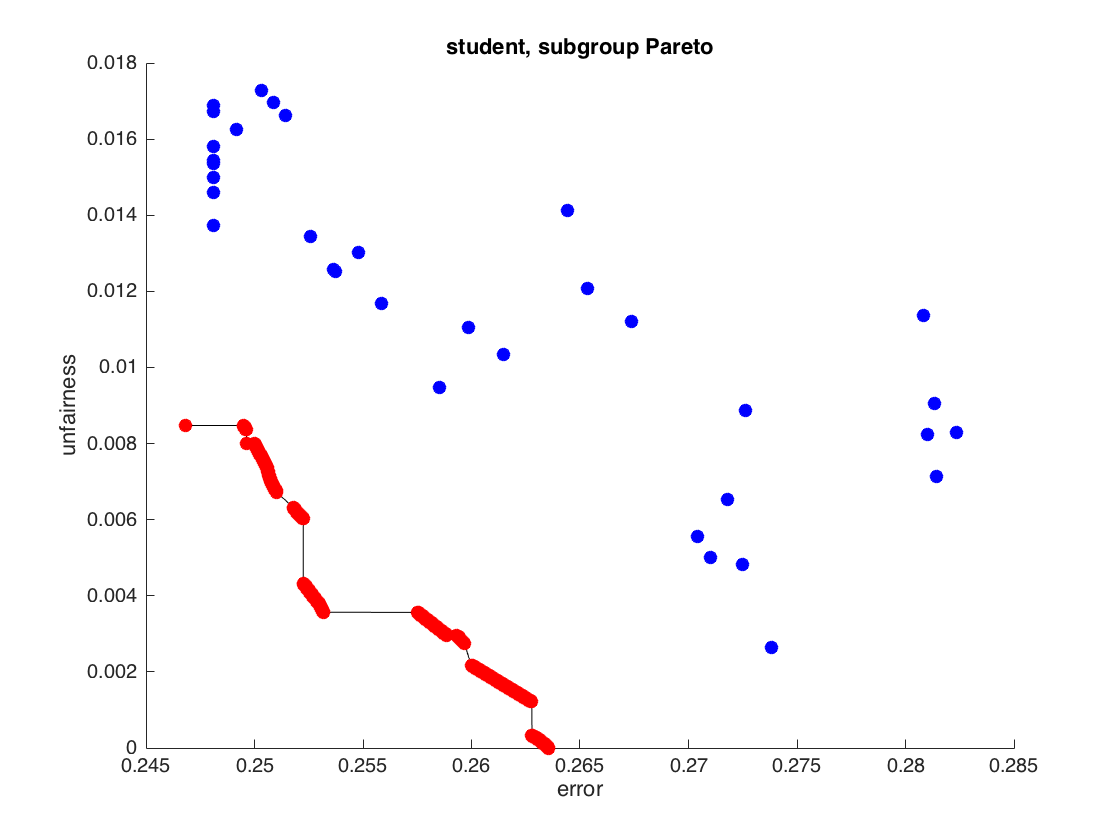}}
\subfigure[]{\includegraphics[scale=0.14]{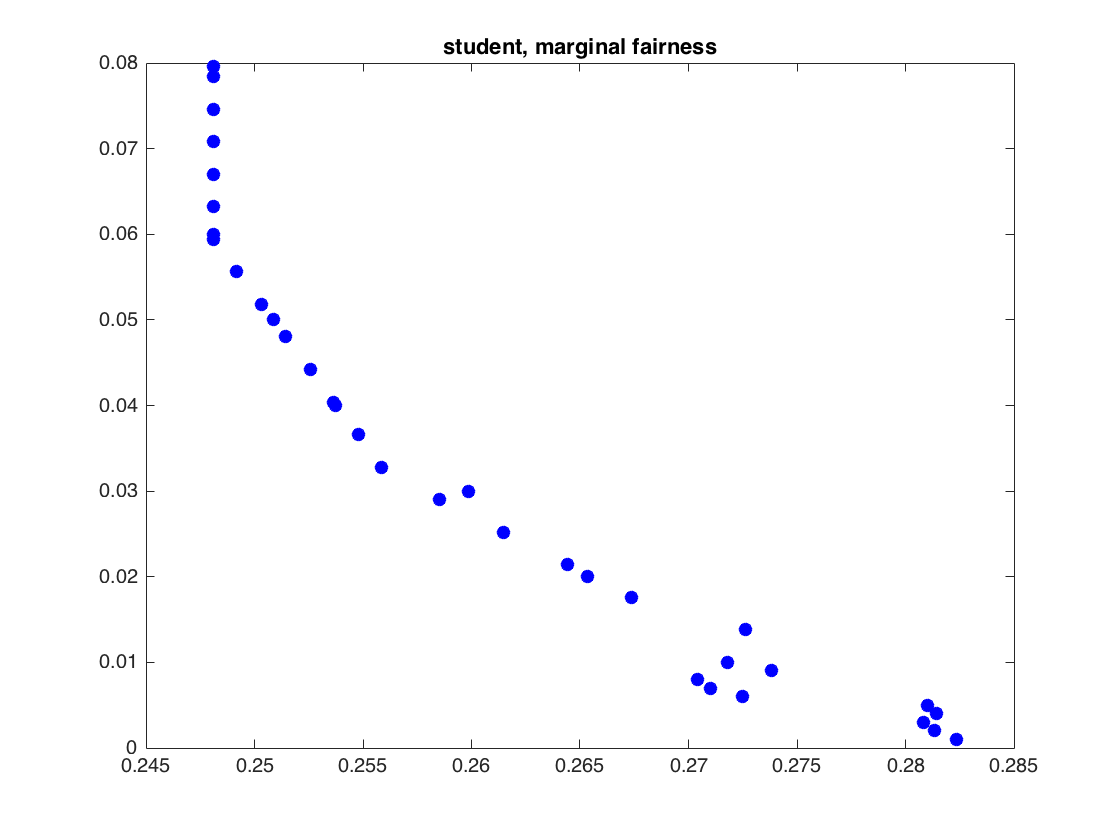}}
\caption{{Left column: The red points show the Pareto frontier of error (x axis) and subgroup fairness violation (y axis)
	for the $\knrw$ algorithm across all four data sets, while the blue points show the error and subgroup fairness violation
	for the models achieved by the $\MSR$ algorithm.
	Right column: The error and marginal fairness violation for the $\MSR$ algorithm across all four data sets.
	Ordering of datasets is Communities and Crime, Law School, Adult, and Student.}}
\label{ccpar}
\end{figure*}

Regardless of convergence, for plots such as those in Figure~\ref{fig:conv}, it is natural to take the
$\langle \epsilon_t, \gamma_t \rangle$
pairs across all $t$ and all input $\gamma$, and compute the undominated or Pareto frontier of these pairs.
This frontier represents the accuracy-fairness tradeoff achieved by the $\knrw$ algorithm on a given data set, which
is arguably its most important output. The choice of where one wants to be on the frontier is a policy question that should be made by
domain experts and stakeholders, and dependent on the stakes involved (e.g. online advertising vs. criminal sentencing).

It is also of interest to compare the subgroup fairness achieved by
the $\knrw$ algorithm (which is explicitly optimizing under a subgroup fairness constraint)
with an algorithm only optimizing under weaker and more
traditional marginal fairness constraints. To this end, we also implemented a version of the algorithm from \cite{MSR} --- which we will refer to as the $\MSR$ algorithm ---
for marginal
fairness.\footnote{Since some of the protected attributes are continuous rather than discrete, and the $\MSR$ algorithm only
handles discrete attributes, in order to run the marginal fairness algorithm we
create sensitive groups by thresholding on the mean of each sensitive attribute.}
From a theoretical perspective, {\em a priori\/}
we would expect models trained for marginal fairness to fare poorly on subgroup fairness. But it
is an empirical question --- perhaps on some datasets, demanding marginal fairness
already suffices to enforce subgroup fairness as well. Thus the high-level
question is whether the $\knrw$ framework and algorithm are worth the added analytical and computational overhead.

In the left column of Figure~\ref{ccpar}, we show the $\knrw$ algorithm Pareto frontiers for subgroup fairness on all four datasets, and also the pairs
achieved by the $\MSR$ algorithm. In the right column, we also separately show the
marginal fairness frontier achieved by the $\MSR$ algorithm. Before discussing the particulars of each dataset,
we first make the following general observations:
\begin{itemize}
\item For most datasets, the $\knrw$ algorithm yields a Pareto curve that frequently lies well below the straight line
	connecting its endpoints (which we can think of as an empirical form of strong convexity),
	and thus there are non-trivial tradeoffs between accuracy and fairness to consider. On some of these
	curves there are regions of steep descent where subgroup unfairness can be reduced significantly
	with negligible increase in error.
\item While the $\MSR$ algorithm performs well with respect to marginal fairness (right column) as expected, it fares much
	worse than the $\knrw$ algorithm on subgroup fairness for three of the datasets. Thus marginal fairness
	is not just theoretically, but also empirically a weaker notion, and generally will not imply subgroup fairness
	``for free''.
\item Nevertheless, there are a handful of points in which the $\MSR$ algorithm produces
	models that actually lie below (and thus dominate) the $\knrw$ Pareto curve by a small amount.
	While this is not possible under the idealized theory --- subgroup fairness is a strictly stronger
	notion than marginal fairness --- it can again be explained by the use of imperfect learning heuristics
	by both algorithms.
\item Focusing just on the $\MSR$ marginal fairness curves in the right column, we see that each of them
	begins with a steep drop, meaning that in every case, the marginal unfairness of the unconstrained
	error-optimal model can be significantly improved with little or no increase in error.
\item By matching points between the $\MSR$ marginal and subgroup fairness plots, we find that
	with the exception of the Student data set, there is a systematic relationship between marginal and
	subgroup unfairness: asking the $\MSR$ algorithm to reduce marginal unfairness also causes it to
	reduce subgroup unfairness --- but not by as much as the $\knrw$ algorithm achieves.
\end{itemize}
Together these observations let us conclude that subgroup fairness is a strong but achievable notion in practice
(at least on these datasets), and that the $\knrw$ algorithm appears to be an effective tool for its investigation.

It is also worth commenting on the differences across datasets, and focusing not just on the qualitative shapes of the
Pareto curves but their actual numerical specifics --- especially since in real applications, these will matter to stakeholders.
For instance, the actual range of error values spanned by the $\knrw$ Pareto curves ranges from nearly 10\% (Communities and Crime)
to less than 2\% (Student). So perhaps for Communities and Crime, the tradeoff is starker from an accuracy perspective. We now provide
some brief commentary on each dataset.\\

\noindent
{\bf Communities and Crime (panels (a) and (b)):}
This is the dataset with perhaps the cleanest and most convex $\knrw$ Pareto curve, with steep drops in subgroup unfairness possible
	for minimal error increase at the beginning.
	In particular are able to reduce the initial $\gamma$-unfairness
	from $0.026$ to less than $0.005$ while only increasing the error from $0.12$ to $0.16$.
	This is a meaningful reduction in unfairness -- e.g. reducing a $26\%$ percent difference
	in false positive rate on a subgroup comprising $10\%$ of the population,
	to a less than $5\%$ false positive rate disparity on a subgroup of the same size.
	Eventually the Pareto curve flattens out, resulting in increasing accuracy costs for reduced unfairness.
	While the $\MSR$ subgroup unfairness curve matches the $\knrw$ Pareto curve on the far left (for all datasets),
	since this corresponds to minimizing error unconstrained by any fairness notion, the outperformance by $\knrw$
	grows rapidly as we make stronger fairness demands.\\

\noindent
{\bf Law School (panels (c) and (d)):}
Here the $\knrw$ Pareto curve appears to be approximately linear, thus providing a constant tradeoff between accuracy and subgroup fairness.
	Interestingly, this is the one dataset in which asking for marginal fairness appears to also yield subgroup fairness for free,
	as the $\MSR$ curve lies very close to the $\knrw$ curve.
	Since this dataset has the fewest number of features overall and the second
	fewest number of protected features, one might be tempted to conjecture that when the
	number of protected features is small, guaranteeing marginal fairness approximately
	guarantees rich subgroup fairness. This claim is falsified by the fact that
	 on the Adult dataset which has similar dimensionality (see below), there is 
	a large gap between the $\knrw$ and $\MSR$ subgroup fairness curves.\\

\noindent
{\bf Adult (panels (e) and (f)):}
Here we see a less smooth $\knrw$ curve, possibly corresponding to the poorer convergence properties on this dataset mentioned earlier. Nevertheless,
	the numerical tradeoff exhibits regions of both steep, inexpensive reduction in unfairness and flat, costly reduction.
	$\MSR$ is again considerably worse when evaluated on subgroup fairness, but still shows a systematic relationship to marginal fairness.\\

\noindent
{\bf Student (panels (g) and (h)):}
Similar to Adult, a varied $\knrw$ curve with multiple tradeoff regimes. This is also the lone dataset in which reducing marginal
	fairness appears to have no relationship to subgroup fairness --- while the $\MSR$ marginal pareto curve in panel (h) remains
	relatively smooth, the subgroup fairness of the corresponding models in panel (d) is now not only worse than for $\knrw$,
	but shows no monotonicity.
	$\knrw$ is able to decrease $\gamma$-unfairness to $0$ with only a $2\%$ increase in error,
	while the $\MSR$ algorithm only drives the subgroup unfairness to $0.002$ at its best,
	with an over $3\%$ increase in error from the unconstrained classifier.\\

Having established the efficacy of subgroup fairness and the $\knrw$ algorithm on the four datasets,
we now turn to experiments and visualizations allowing us to better understand the behavior and dynamics
of the algorithm.

\subsection{Flattening the Discrimination Surface}
\label{subsec:heatmap}

\begin{figure*}[p]
\subfigure[]{\includegraphics[scale=0.15]{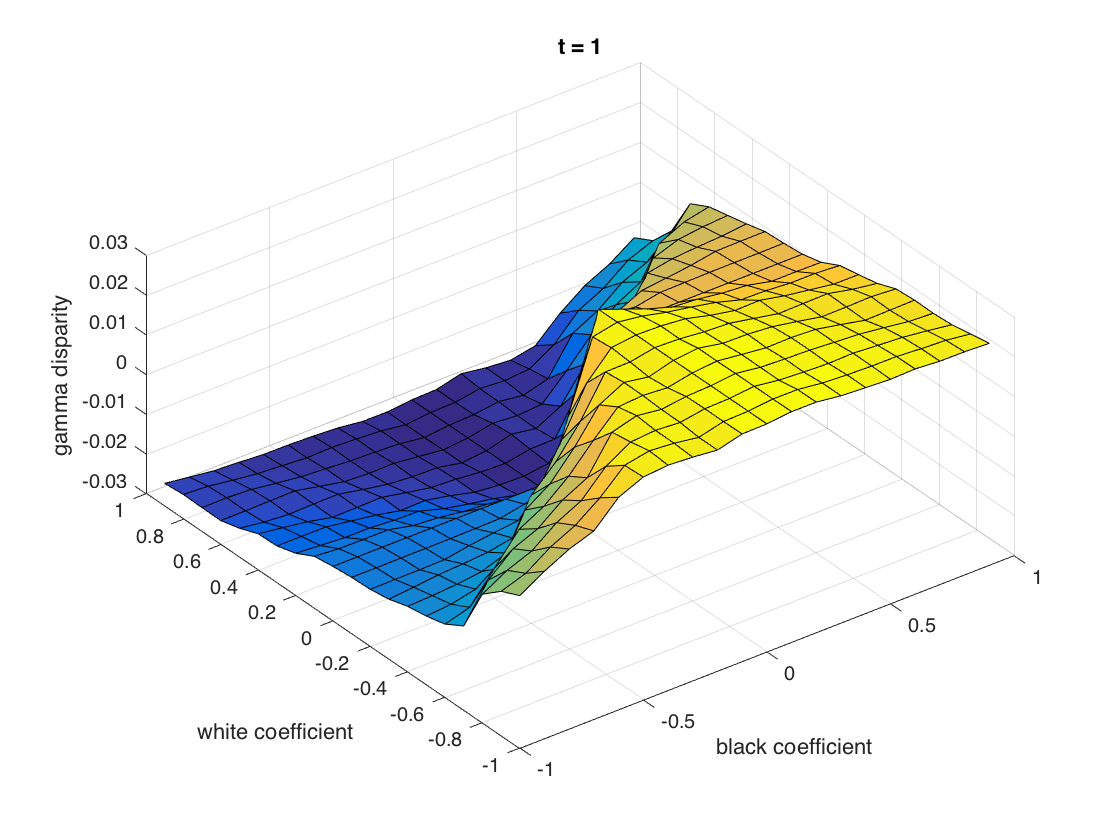}}
\subfigure[]{\includegraphics[scale=0.15]{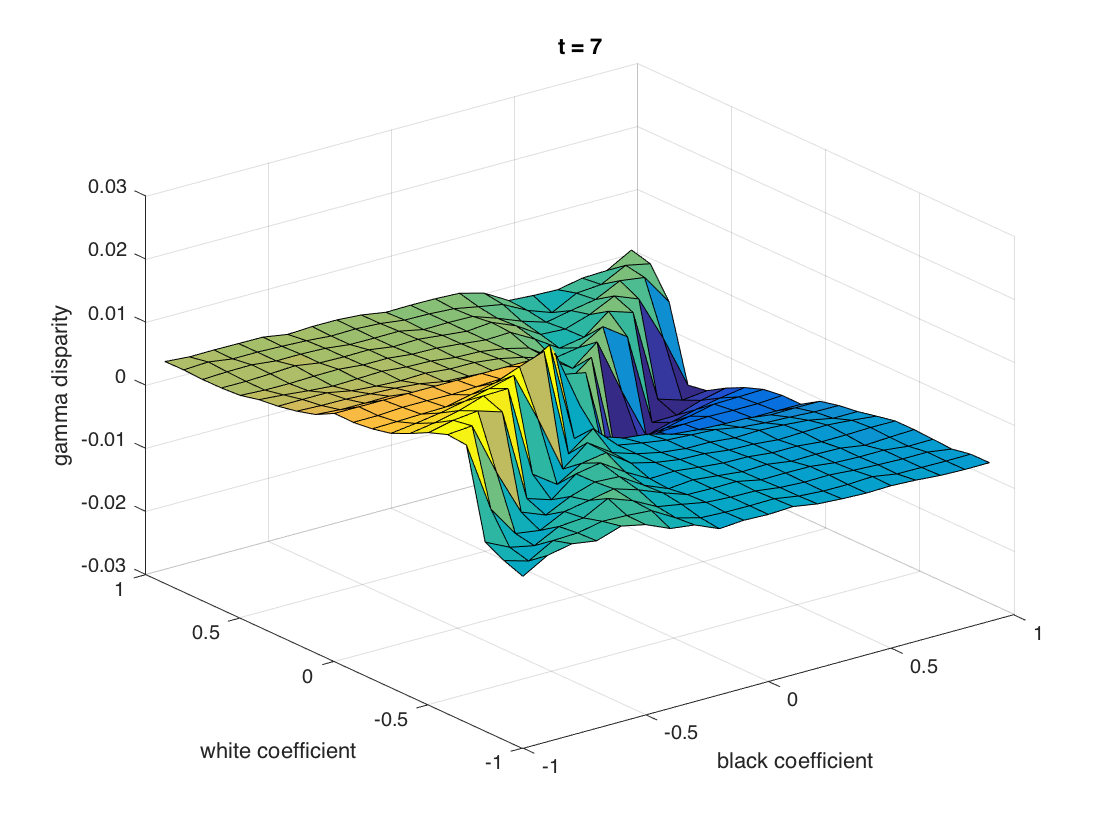}}
\subfigure[]{\includegraphics[scale=0.15]{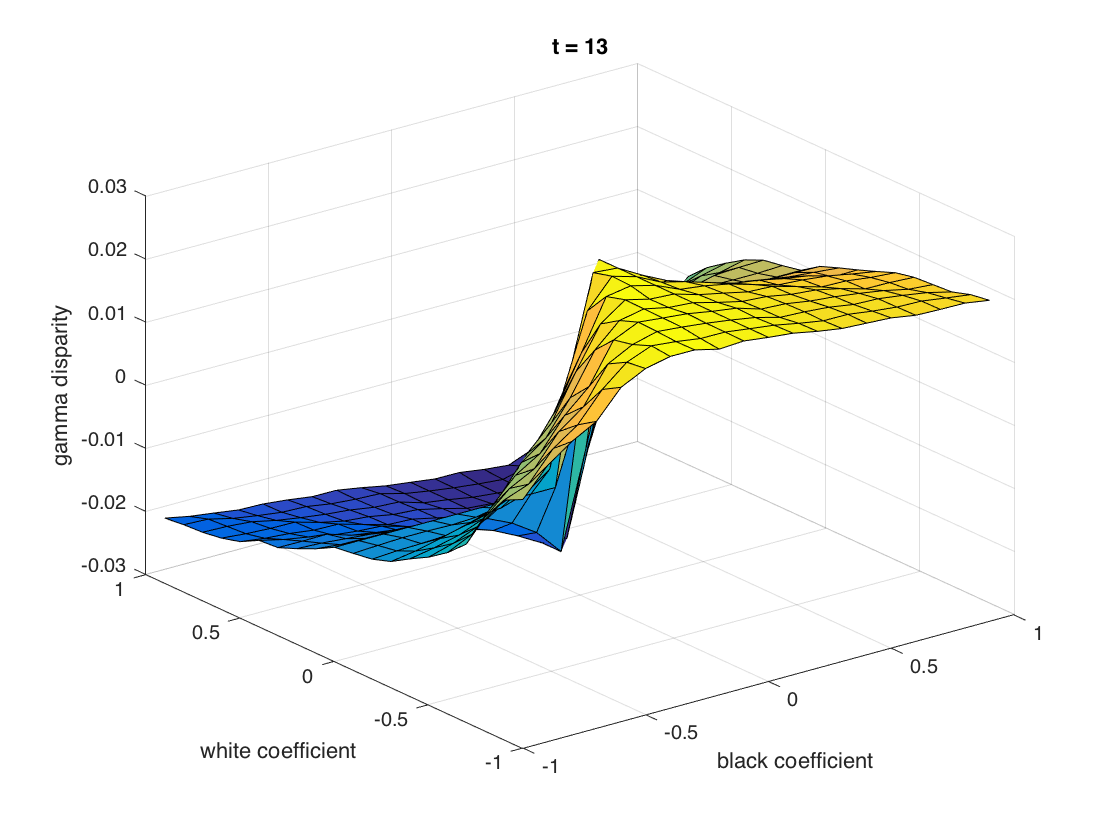}}
\subfigure[]{\includegraphics[scale=0.15]{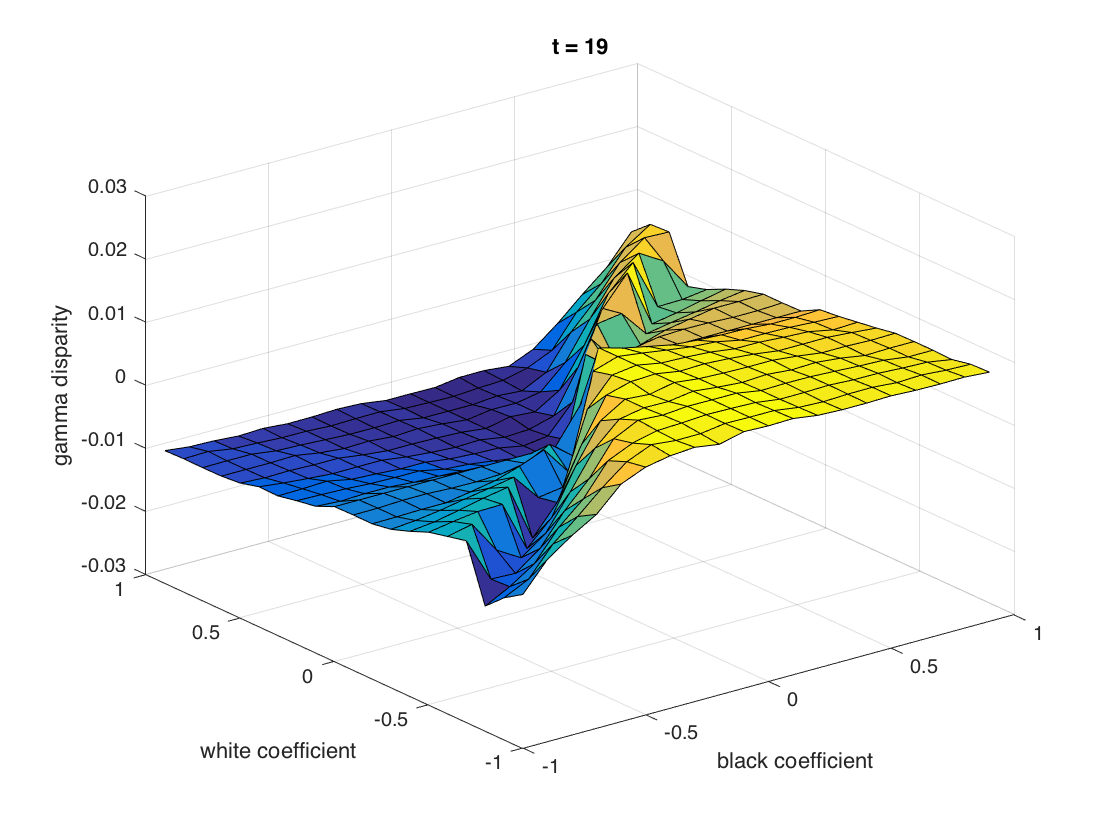}}
\subfigure[]{\includegraphics[scale=0.15]{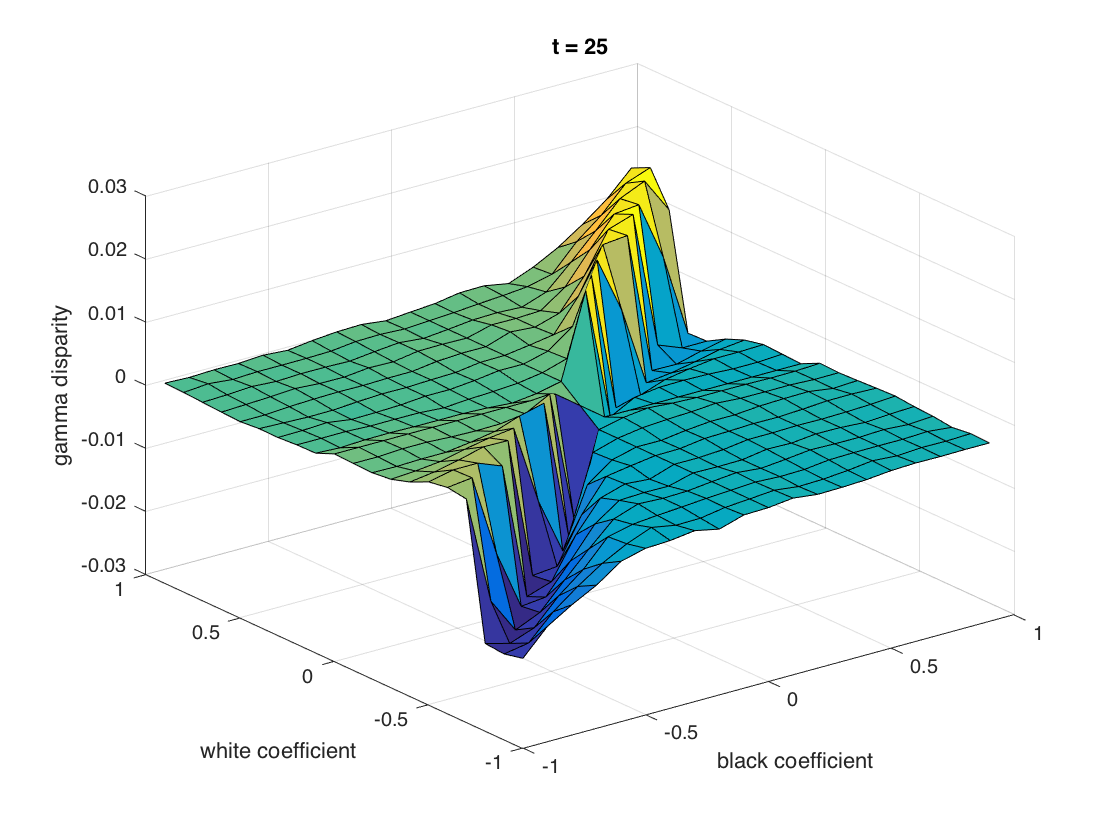}}
\subfigure[]{\includegraphics[scale=0.15]{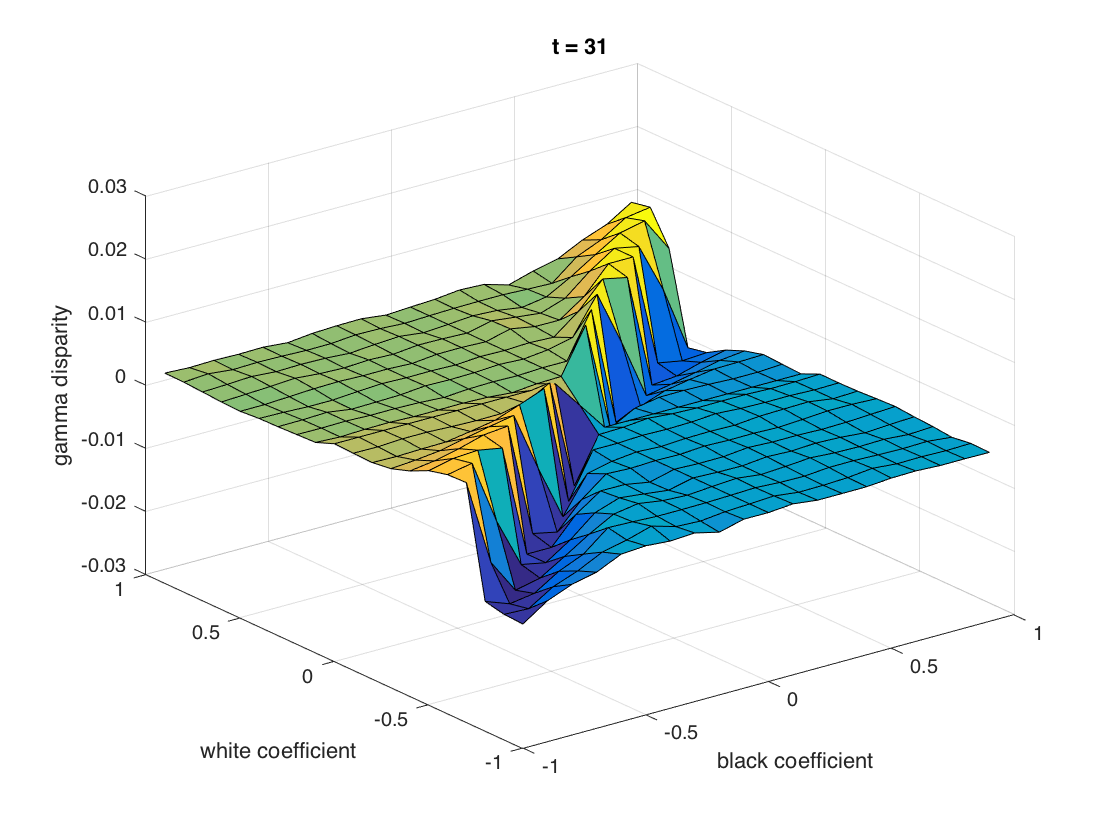}}
\subfigure[]{\includegraphics[scale=0.15]{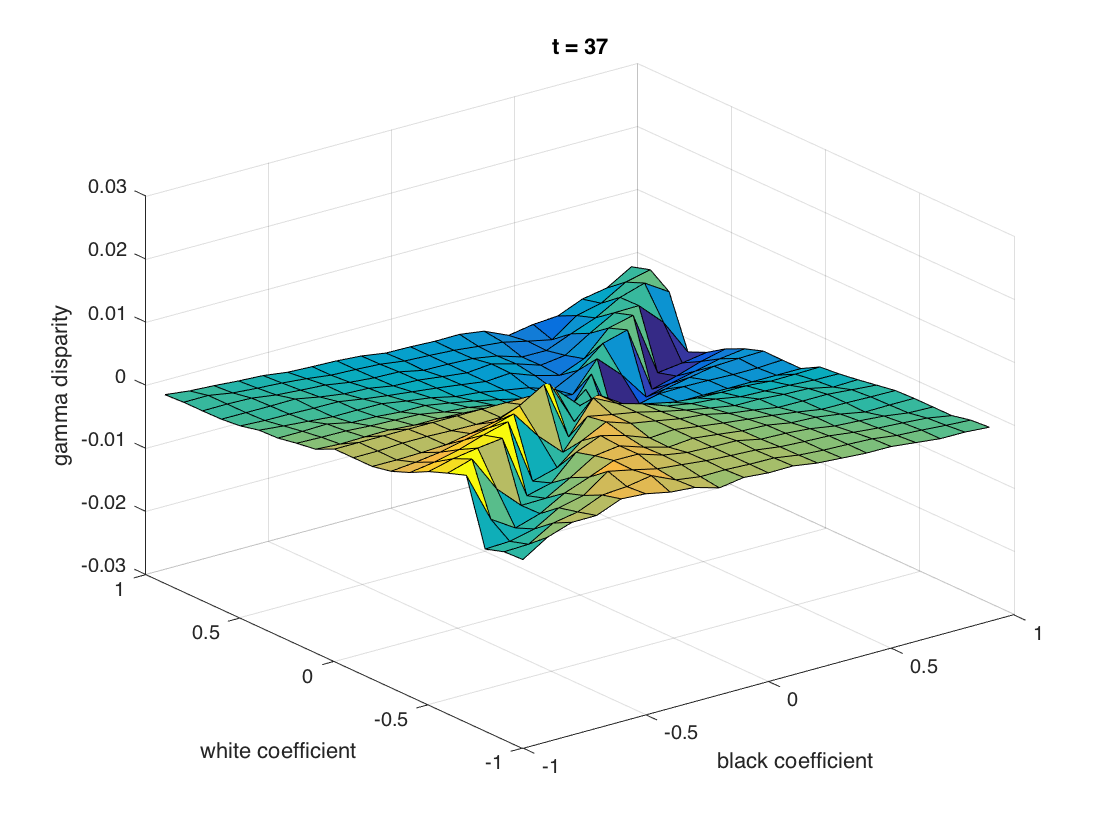}}
\hspace{1.75in}
\subfigure[]{\includegraphics[scale=0.15]{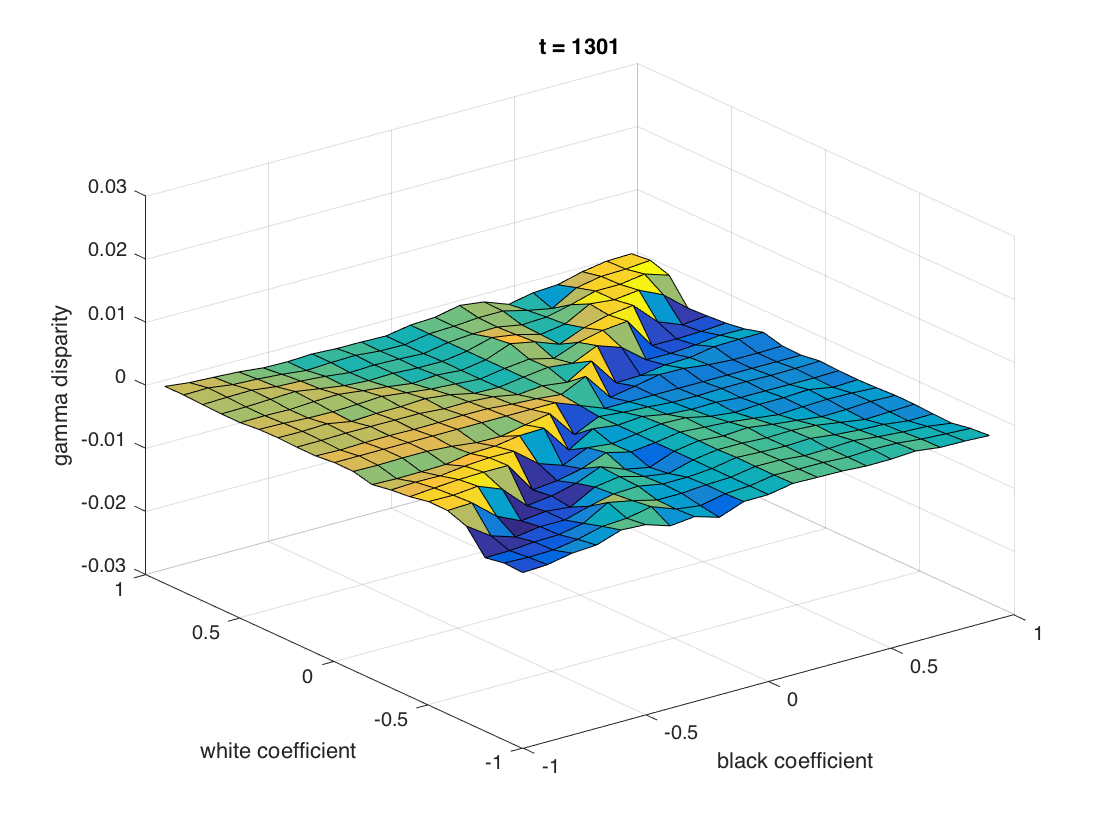}}
\caption{
Evolution of discrimination surface for the $\knrw$ algorithm from $t = 1 \ldots 1301$.
Each point in the plane corresponds to a different subgroup over two protected attributes,
and the corresponding $z$ value is the current false positive discrepancy for the subgroup.
}
\label{fig:heatmap}
\end{figure*}

Recall that in the various analyses and plots above, we rely on the
Auditor of $\knrw$ to detect unfairness. This Auditor is in turn a heuristic,
relying on an optimization procedure without any theoretical guarantees, which could potentially fail in practice.
This means that while any detected unfairness is a lower bound on the true subgroup unfairness,
it could be the case that the heuristic Auditor is simply failing to detect a larger disparity,
and that the models learned by $\knrw$ look more fair than they really are.

We explore this possibility on the Communities \& Crime dataset by implementing a brute force
Auditor that runs alongside the $\knrw$ algorithm. To make brute force auditing computationally tractable,
we designate only two attributes as protected; \textit{pctwhite} and \textit{pctblack},
the percentage of each community that consists of white and black people respectively.
While the $\knrw$ algorithm uses the same heuristic Auditor it always does,
at each round we also perform a brute force audit as follows.
Subgroups $g_\theta$ are defined by a linear threshold function $\theta$ over the $2$ sensitive attributes, e.g. $(x_1, x_2) \in g_\theta$ iff
$\langle \theta, (x_1, x_2) \rangle \geq 0$. We discretize $\theta \in [-1,1]^2$ in increments of $0.1$, and for the subgroup defined by each $\theta$ in the discretization we compute the $\gamma$-unfairness. Hence at each round we can take the current classifier of the Learner, and plot for each group $g_\theta$ the point $(\theta_1, \theta_2, \gamma)$.

Note that in addition to making brute force auditing tractable, restricting to two dimensions permits direct visualization of discrimination.
In Figure~\ref{fig:heatmap}, we show a sequence of ``discrimination surfaces'' for the $\knrw$ algorithm over the $2$ protected features,
with input $\gamma = 0$. The $x-y$ axes are the coefficients of $\theta$
corresponding to \textit{whitepct} and \textit{blackpct} respectively, and the $z$-axis is the $\gamma$-unfairness of
the corresponding subgroup. This is our first non-heuristic view of $\gamma$-unfairness, and
also shows us the entire surface of $\gamma$-unfairness, rather than just the most violated subgroup.
Note that perfect subgroup fairness would correspond to an entirely flat discrimination surface at $z = 0$.

We observe first that the unconstrained classifier in $t = 1$ (panel (a))
shows a very systematic bias along the lines of our sensitive attributes. In particular groups with \textit{whitepct} $> 0$ and \textit{blackpct} $< 0$, e.g. communities with large numbers of white residents and relatively fewer black residents have a much higher false positive rate for being classified as violent.
Conversely, majority black communities are less likely to be incorrectly labeled as violent.
The mean $\gamma$-unfairness (base rate - community rate) for  \textit{whitepct} $> 0$,  \textit{blackpct} $< 0$
communities is $-0.0242$, whereas the mean for \textit{whitepct} $< 0$,  \textit{blackpct} $ > 0$ groups is
$0.0247$. The maximum $\gamma$-unfairness in $t = 1$ is $0.028$, and $61.25\%$ of the $400$ subgroups have $\gamma$-unfairness $ > 0.02$. Recall that this corresponds to e.g. a $20\%$ disparity of the false positive rate from the base rate, for groups as large as 10\% of the population. 
We are thus far from perfect subgroup fairness.

As the algorithm proceeds, we see this discrimination flip by $t = 7$ (panel (b)),
into a regime with a higher false positive rate for predominantly black communities,
and then revert again by $t = 13$. Over the early iterations these oscillations continue, growing less
drastic as the $\gamma$-unfairness surface starts to flatten out noticeably by $t = 37$ (panel (g)).
In panel (h) we plot $t = 1301$ and see that the surface has almost completely
flattened, with maximum $\gamma$-unfairness below $.0028$.
So over the course of the first $1300$ iterations of $\knrw$ we've reduced the
$\gamma$-unfairness from over $0.02$ in most of the subgroups,
to less than $0.0028$ in \textit{every} subgroup. 
Recall again that this corresponds to false positive rate disparities of at most 2.8\% in subgroups that represent 10\% of the population --- a reduction from false positive rate disparities of 20\% many similarly sized subgroups. This represents an order of magnitude improvement that results from using the classifier learned by $\knrw$. 

\subsection{Understanding the Dynamics}
\label{subsec:traj}

\begin{figure*}[h]
\centering
\subfigure[]{\includegraphics[scale=0.15]{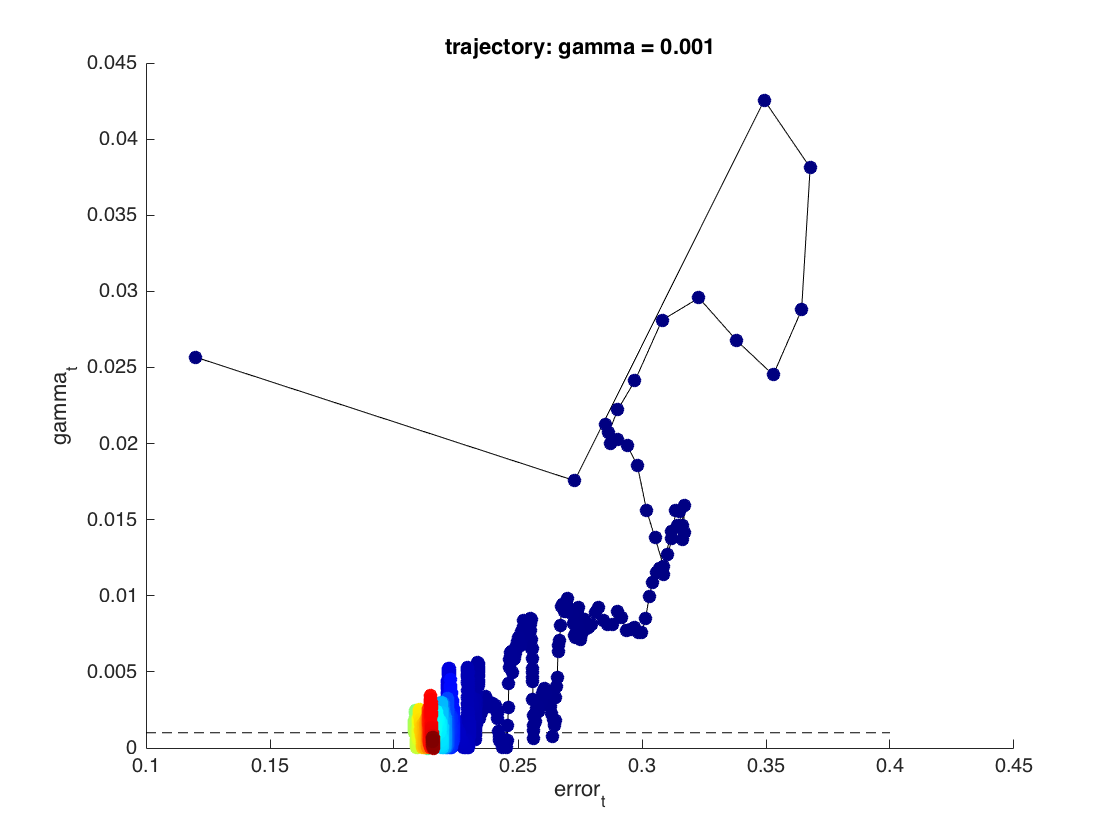}}
\subfigure[]{\includegraphics[scale=0.15]{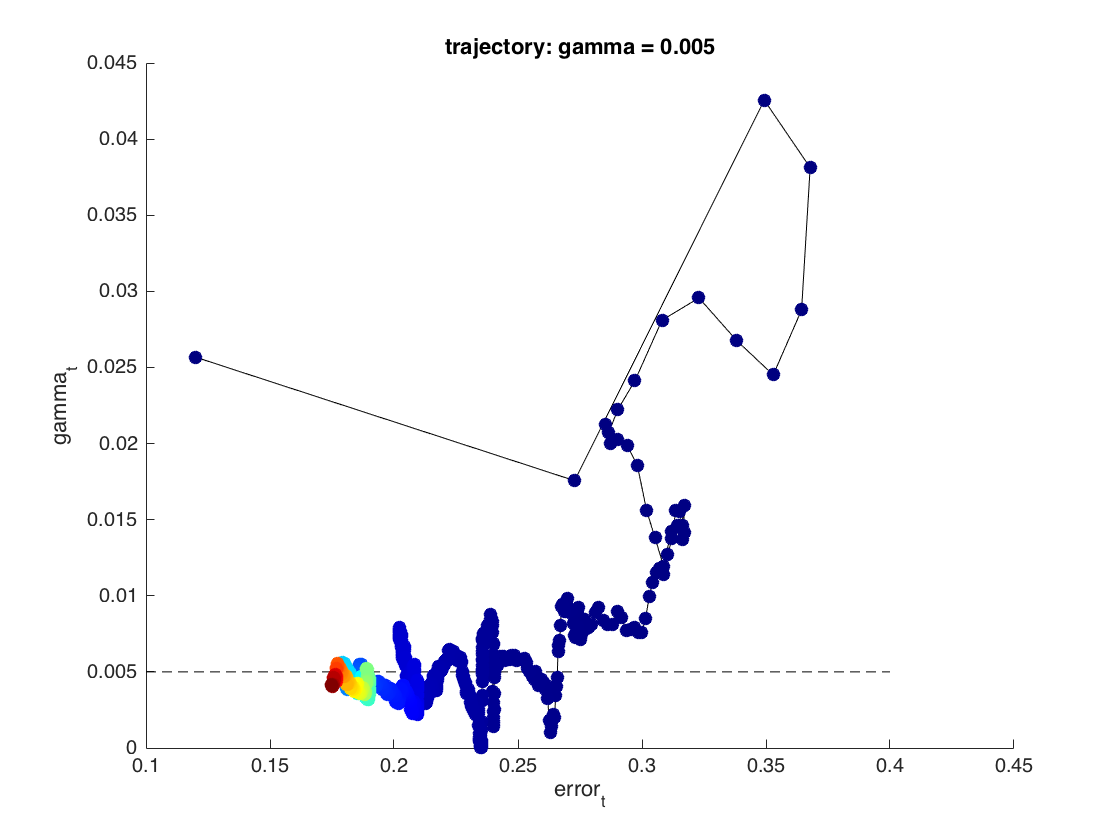}}
\subfigure[]{\includegraphics[scale=0.15]{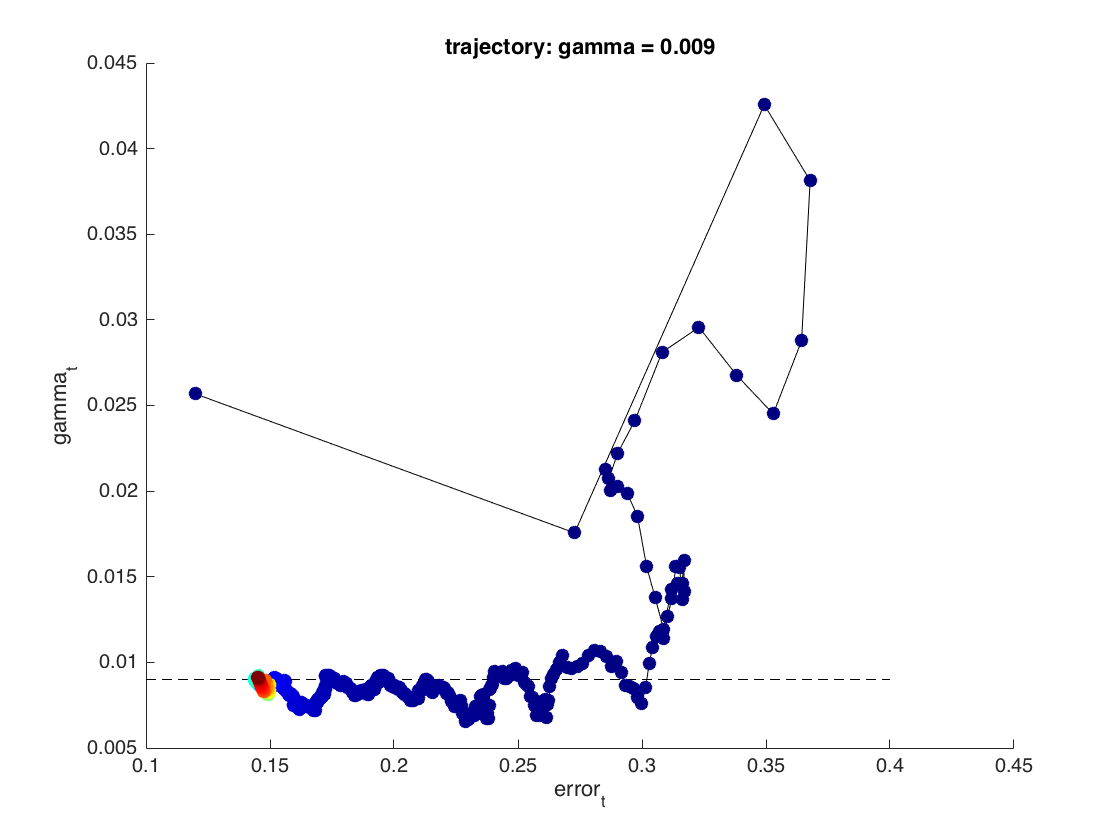}}
\subfigure[]{\includegraphics[scale=0.15]{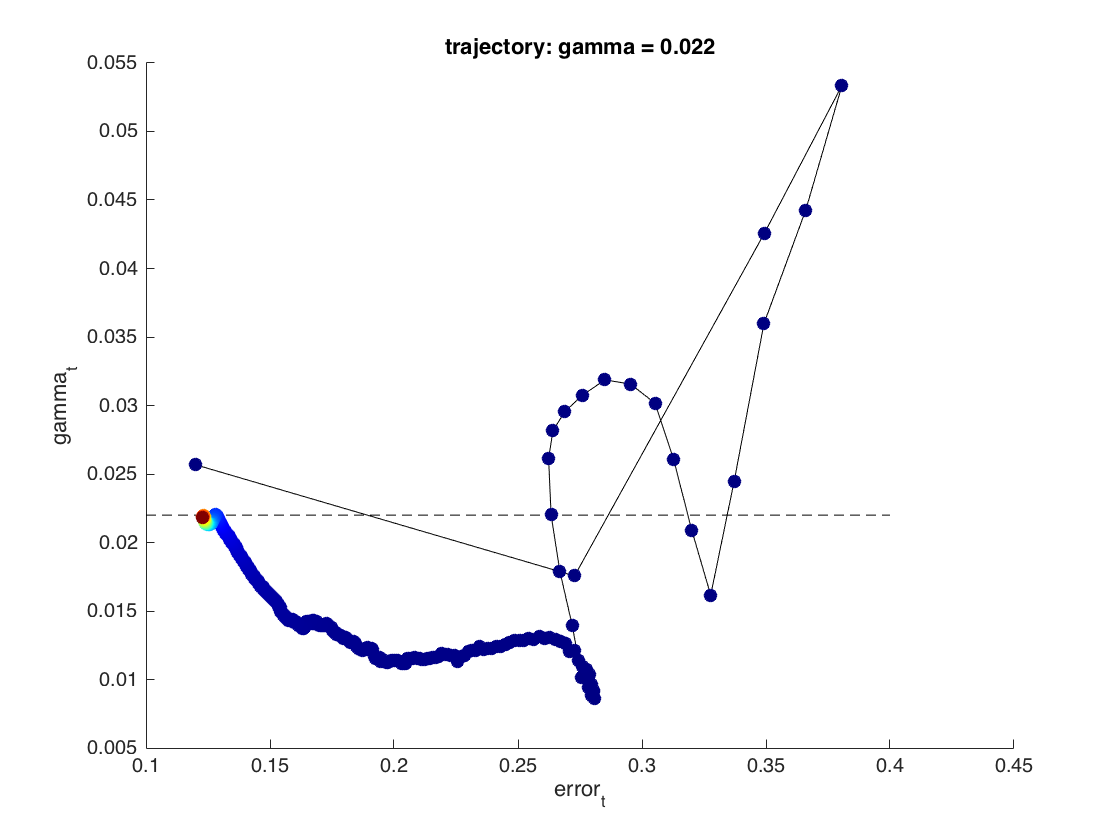}}
\caption{
$\langle \epsilon_t, \gamma_t \rangle$ trajectories for
Communities and Crime, for $\gamma \in \{0.001, 0.005, 0.009, 0.022\}.$
}
\label{traj}
\end{figure*}

We conclude by examining the dynamics of the $\knrw$ algorithm on the Communities and Crime dataset in greater detail.
More specifically, since the algorithm is formulated as a game between a Learner who at each iteration $t$ is trying to minimize the
error $\epsilon_t$, and an Auditor who is trying to minimize subgroup unfairness $\gamma_t$, we visualize the trajectories
traced in $\langle \epsilon_t, \gamma_t \rangle$ space as $t$ increases.

The plots in Figure~\ref{traj} correspond to such trajectories for input $\gamma$ values of $0.001, 0.005, 0.009,$ and $0.022$
(panels (a), (b), (c) and (d) respectively), which are
denoted by the dashed lines on the $\gamma_t$ axis of each figure. The $0.001$ and $0.005,0.009$ values correspond to small
and intermediate $\gamma$ regimes, whereas $0.022$ is close to (but slightly below)
the subgroup unfairness of the unconstrained classifier. The trajectories are color coded from colder
to warmer colors according to their iteration number to give a sense of speed of convergence.

The first plot in all four trajectories corresponds to the
$\langle \epsilon_0, \gamma_0 \rangle$
of the unconstrained classifier. Furthermore, as long as the current $\gamma_t$ values remain above the
horizontal dashed line representing the input $\gamma$, the trajectories remain identical, as the same subgroups are
being presented to the learner in each trajectory. But when $\gamma_t$ falls below a given input $\gamma$, that
trajectory will follow its own path going forward.

We first observe that the dynamics exhibit a fair amount of complexity and subtlety. They all begin with low error and
large unfairness, and quickly follow a brief but large increase in $\epsilon_t$ as fairness starts to be enforced. There
are steps in which both $\epsilon_t$ and $\gamma_t$ increase, and a large early loop in trajectory space is observed.
But the first three trajectories (panels (a), (b) and (c), corresponding to the three smaller values of $\gamma$) quickly settle near the
input $\gamma$ line, at which point begins a long, oscillatory ``border war'' around this line, as the Learner tries to minimize
error, but is pushed back below the line by the Auditor anytime $\gamma$-fairness is violated. The idealized theory predicts
that each trajectory should end at the input $\gamma$ line (subgroup fairness constraint saturated), and with larger
input $\gamma$ (weaker fairness constraint) resulting in lower error. The empirical trajectories indeed conform nicely
to the theory, with the final (red) points near the dashed lines, and further left for larger $\gamma$.

Panel (d), corresponding to a much larger input $\gamma$, diverges much earlier from the other three (on its second step),
and early on sees unfairness driven far below the specified value. The dynamics then see a slow, gradual decrease of
error and increase of unfairness back to the input value, with the trajectory ending up near where it began, but just
slightly more fair, as specified by $\gamma$.


%% file: conclusions.tex
\section{Conclusions}

In this work we have established the empirical efficacy of the
notion of rich subgroup fairness and the algorithm of
\cite{KNRW18} on four fairness-sensitive datasets, and the necessity of
explicitly enforcing subgroup (as opposed to only marginal) fairness.
There are
a number of interesting directions for further experimental
work we plan to pursue, including:
\begin{itemize}
\item Experiments with richer Learner model classes $\cH$, while
	keeping the Auditor subgroup class $\cF$ relatively simple
	and fixed. One conjecture is that by making the hypothesis
	space richer, more appealing Pareto curves may be achieved.
	There is also some rationale for keeping $\cF$ simple, since
	we would like to have some intuitive interpretation of what
	the subgroups represent, while the same constraint may not
	hold for $\cH$.
\item Implementation and experimentation with the no-regret algorithm
	of \cite{KNRW18}, which may have superior convergence and other
	properties due to its stronger theoretical guarantees.
\item Experiments on the generalization performance of subgroup
	fairness in the form of test-set Pareto curves. While as mentioned,
	standard VC theory can be applied to obtain worst-case bounds, one
	might expect even better empirical generalization.
\end{itemize}

%% file: appendix.tex
\section{Details of the $\knrw$ algorithm}
We here recall the main results of \cite{KNRW18}.  Let $S$ denote a
set of $n$ labeled examples $\{z_i = (x_i, x'_i), y_i)\}_{i=1}^n$, and
let $\cP$ denote the empirical distribution over this set of examples.
Let $D$ denote a probability distribution over $\cH$. Consider the
following {\em Fair ERM (Empirical Risk Minimization)} problem:
\begin{align}
  &  \min_{D\in \Delta_\cH}\; \Ex{h\sim D}{err(h, \cP)}\\
  \mbox{such that } \forall g\in \cF \qquad &
\fpsize(g, \cP) \cdot \fpdisp(g, D, \cP) \leq \gamma.
\end{align}
where $err(h, \cP) = \Pr_\cP[h(x, x') \neq y]$, and the quantities
$\fpsize$ and $\fpdisp$ are defined in \Cref{fp-fair}.
We will write $\OPT$ to denote the objective value at the optimum for
the Fair ERM problem, that is the minimum error achieved by a
$\gamma$-fair distribution over the class $\cH$. This is the fair learning problem that we want to solve.   We assume our algorithm has access to the cost-sensitive
  classication oracles $\CSC(\cH)$ and $\CSC(\cF)$ over the classes
  $\cH$ and $\cF$ respectively.

Kearns et al. \cite{KNRW18} prove the existence of an oracle-efficient algorithm for solving the Fair ERM problem:

\begin{theorem}[\cite{KNRW18}]\label{thm:polytime}
  Fix any $\nu, \delta\in (0, 1)$. Then given an input of $n$ data
  points and accuracy parameters $\nu, \delta$ and access to oracles
  $\CSC(\cH)$ and $\CSC(\cF)$, there exists an algorithm that runs in
  polynomial time, and with probability at least $1 - \delta$, outputs
  a randomized classifier $\hat D$ such that
  $err(\hat D, \cP) \leq \OPT + \nu$, and for any $g\in \cF$, the
  fairness constraint violations satisfies
  \[
    \alpha_{FP}(g, \cP) \cdot\beta_{FP}(g, \hat D, \cP) \leq \gamma + O(\nu).
  \]
\end{theorem}

The algorithm corresponding to Theorem \ref{thm:polytime} is randomized, however, and hence less amenable to the sort of empirical investigation that we undertake in this paper. Fortunately, \cite{KNRW18} also give another algorithm, with somewhat weaker guarantees. It has the same guarantees as Theorem \label{thm:polytime}, except the convergence guarantees hold only after an exponential, rather than a polynomial number of steps. However, it has the virtue of very simple per-step dynamics, and is the algorithm that we investigate in this paper. Its pseudo-code follows:

\begin{algorithm}[h]
  \caption{\bf{FairFictPlay: Fair Fictitious Play}}
 \label{alg:fairfict}
  \begin{algorithmic}
    \STATE{\textbf{Input:} distribution $\cP$ over the labelled data
      points, CSC oracles $\CSC(\cH)$ and $\CSC(\cG)$ for the classes
      $\cH(S)$ and $\cG(S)$ respectively, dual bound $C$, and number
      of rounds $T$}

    \STATE{\textbf{Initialize}: set $h^0$ to be some classifier in
      $\cH$, set $\lambda^0$ to be the zero vector. Let
      $\overline D$ and $\overline \lambda$ be the point distributions that put
      all their mass on $h^0$ and $\lambda^0$ respectively.}

    \STATE{\textbf{For} $t = 1, \ldots, T$:}

    \INDSTATE{\textbf{Compute the empirical play distributions}:}

    \INDSTATE[2]{\textbf{Let} $\overline D$ be the uniform distribution
      over the set of classifiers $\{h^0, \ldots, h^{t-1}\}$ }

    \INDSTATE[2]{\textbf{Let}
      $\overline \lambda = \frac{\sum_{t'< t} \lambda^{t'}}{t}$ be the
      auditor's empirical dual vector}

    \INDSTATE{\textbf{Learner best responds}:
      Use the oracle $\CSC(\cH)$ to compute
      $h^{t} = \argmin_{h \in \cH(S)} \langle \LC(\overline\lambda), h
      \rangle$
}

\INDSTATE{\textbf{Auditor best responds}: Use the oracle $\CSC(\cF)$
  to compute
  $\lambda^t = \argmax_{\lambda} \Ex{h\sim \overline D}{U(h,
    \lambda)}$
}

    \STATE{\textbf{Output:} the final empirical distribution
      $\overline D$ over classifiers}

    \end{algorithmic}
  \end{algorithm} 
  
  To briefly introduce the notations in the description above, we
  first note that we can rewrite the set of constraints in the Fair
  ERM problem as follows: for each $g\in \cG(S)$, 
\begin{align}
    &\Phi_+(h, g) \equiv \fpsize(g, P) \, \left(\FP(h) - \FP(h, g) \right) - \gamma\leq 0 \label{f1}\\
  &  \Phi_-(h, g)\equiv \fpsize(g, \cP)\, \left(\FP(h, g) - \FP(h)\right) - \gamma \leq 0 \label{f2}
\end{align}
Here $\cG(S)$ and $\cH(S)$ denote the set of all labellings on $S$
that are induced by $\cG$ and $\cH$ respectively, that is
\begin{align}
  &\cG(S) = \{(g(x_1), \ldots , g(x_n)) \mid g\in \cG\} \qquad
  \mbox{and,} \\
&\cH(S) = \{(h(X_1), \ldots , h(X_n))\mid h\in
  \cH\}
\end{align}
Then, $\lambda$ is a vector with a coordinate $\lambda_g^+$ and
$\lambda_g^-$ for every subgroup $g \in \cF$ such that $\lambda_g^+$
and $\lambda_g^-$ are dual variables that corresponds the pair of
constraints \eqref{f1} and \eqref{f2}.  The partial Lagrangian of the
linear program is the following:
\begin{align*}
  \cL(D, \lambda) = \Ex{h\sim D}{err(h, \cP)} +  \sum_{g\in \cG(S)}
  \left( \lambda_g^+\,  \Phi_+(D, g) +  \lambda_g^- \,  \Phi_-(D,g) \right)
\end{align*}
Similarly, the payoff function for the zero-sum game is then defined
as: for any pair of actions
$(h, \lambda)\in \cH\times \Lambda_{\text{pure}}$,
\[
  U(h, \lambda) = err(h, \cP) + \sum_{g\in \cG(S)}
  \left(\lambda_g^+\Phi_+(h, g) + \lambda_g^-\Phi_-(h, g) \right).
\]
Given a fixed set of dual variables $\lambda$, we will write
$\LC(\lambda) \in \RR^n$ to denote the vector of costs for labelling
each datapoint as $1$. That is, $\LC(\lambda)$ is the vector such that
for any $i \in [n]$, $\LC(\lambda)_i = c_i^1$.

We can find a best response for the Learner by making a call to the
cost-sensitive classification oracle. In particular, we assign costs
to each example $(X_i, y_i)$ as follows:
\begin{itemize}
\item if $y_i = 1$, then $c_i^0 = 0$ and $c_i^1 = - \frac{1}{n}$;
\item otherwise, $c_i^0 = 0$ and
  \begin{align}
    c_i^1 = \frac{1}{n} &+ \frac{1}{n}\sum_{g\in \cG(S)}
                          (\lambda_g^+ - \lambda_g^-) \left(\Pr[g(x) = 1\mid y = 0] -  \mathbf{1}[g(x_i) = 1]\right)
  \end{align}
\end{itemize}
Then given a fixed set of dual variables $\lambda$, we will write
$\LC(\lambda) \in \RR^n$ to denote the vector of costs for labelling
each datapoint as $1$. That is, $\LC(\lambda)$ is the vector such that
for any $i \in [n]$, $\LC(\lambda)_i = c_i^1$.